\documentclass[conference]{IEEEtran}

\usepackage[]{hyperref}
\usepackage[dvipsnames]{xcolor}
\usepackage[normalem]{ulem}
\usepackage{graphicx}
\usepackage{amsmath,amsfonts,amssymb,amstext,mathtools,bm,amsthm}
\usepackage{booktabs}
\usepackage{outlines}
\usepackage{macros}
\usepackage{tablefootnote}
\usepackage{makecell}
\usepackage{multicol}
\usepackage{multirow}
\usepackage{stackengine}
\usepackage{enumitem}
\usepackage{wrapfig}
\usepackage{tikz,pgfplots}

\usepackage[framemethod=TikZ]{mdframed}
\mdfsetup{skipabove=5pt,
innertopmargin=-5pt}

\usepackage{makecell}

\usepackage[labelfont=bf,font=footnotesize]{caption}
\usepackage[labelfont=bf,font=footnotesize]{subcaption}
\usepackage[numbers,sort&compress]{natbib}

\usepackage[ruled]{algorithm}
\usepackage{algorithmicx}
\usepackage[noend]{algpseudocode}

\newcommand{\revise}[1]{#1}
\newcommand{\remove}[1]{}

\makeatletter
\newcommand{\printfnsymbol}[1]{%
  \textsuperscript{\@fnsymbol{#1}}%
}
\makeatother

\begin{document}

\title{Toward Certified Robustness Against Real-World Distribution Shifts}

\DeclareRobustCommand*{\IEEEauthorrefmark}[1]{%
  \raisebox{0pt}[0pt][0pt]{\textsuperscript{#1}}%
}

\author{
\IEEEauthorblockN{
Haoze Wu\textsuperscript{\textsection}\IEEEauthorrefmark{1},
Teruhiro Tagomori\textsuperscript{\textsection}\IEEEauthorrefmark{1,2},
Alexander Robey\textsuperscript{\textsection}\IEEEauthorrefmark{3},
Fengjun Yang\textsuperscript{\textsection}\IEEEauthorrefmark{3},
Nikolai Matni\IEEEauthorrefmark{3},\\
George Pappas\IEEEauthorrefmark{3},
Hamed Hassani\IEEEauthorrefmark{3},
Corina Pasareanu\IEEEauthorrefmark{4},
Clark Barrett\IEEEauthorrefmark{1}
}
\IEEEauthorblockA{
\IEEEauthorrefmark{1} Stanford University, USA.
\IEEEauthorrefmark{2} NRI Secure, Japan.
\IEEEauthorrefmark{3} University of Pennsylvania, USA.\\
\IEEEauthorrefmark{4} Carnegie Mellon University, USA.}
\IEEEauthorblockA{
\IEEEauthorrefmark{1} \{haozewu,barrett\}@cs.stanford.edu
\IEEEauthorrefmark{2} tteruhiro@nri-secure.com \\
\IEEEauthorrefmark{3} \{arobey1,fengjun,nmatni,pappasg,hassani\}@seas.upenn.edu
\IEEEauthorrefmark{4} pcorina@andrew.cmu.edu}
}


\maketitle

\begin{abstract}
We consider the problem of certifying the robustness of deep neural networks against real-world distribution shifts.  To do so, we bridge the gap between hand-crafted specifications and realistic deployment settings by considering a neural-symbolic verification framework in which generative models are trained to learn perturbations from data and specifications are defined with respect to the output of these learned models.  A pervasive challenge arising from this setting is that although \revise{S-shaped activations (e.g., sigmoid, tanh)} are common in the last layer of deep generative models, existing verifiers cannot tightly approximate \revise{S-shaped}\remove{sigmoid} activations.
To address this challenge, we propose a general meta-algorithm for handling \revise{S-shaped}\remove{sigmoid} activations which leverages classical notions of counter-example-guided abstraction refinement. The key idea is to ``lazily'' refine the abstraction of \revise{S-shaped}\remove{sigmoid} functions to exclude spurious counter-examples found in the previous abstraction, thus guaranteeing progress in the verification process while keeping the state-space small.  For networks with sigmoid activations, we show that our technique outperforms state-of-the-art verifiers on certifying robustness against both canonical adversarial perturbations and numerous real-world distribution shifts.  Furthermore, experiments on the MNIST and CIFAR-10 datasets show that distribution-shift-aware algorithms have significantly higher certified robustness against distribution shifts.
\end{abstract}

\section{Introduction}
\label{sec:intro}

Despite remarkable performance in various domains, it is well-known that deep neural networks (DNNs) are susceptible to seemingly innocuous variation in their input data.  
Indeed, recent studies have conclusively shown that DNNs are vulnerable to a diverse array of changes ranging from norm-bounded perturbations~\cite{fgsm,pgd,wong,zhang2019theoretically,kannan2018adversarial,moosavi2016deepfool,robey2021adversarial} to distribution shifts in weather conditions in perception tasks~\cite{robey2020model,wong2020learning,hendrycks2019benchmarking,hendrycks2020many,koh2020wilds}.
To address these concerns, there has been growing interest in using formal methods to obtain rigorous verification guarantees for neural networks with respect to particular specifications~\cite{KaBaDiJuKo17Reluplex,marabou,mipverify,scaling,babsr,xu2020fast,planet,dependency,optAndAbs,peregrinn,mip01,dual,dual1,dual2,soi,ferrari2022complete,nnv,dlv,deeppoly,kpoly,frown,crown,reluval,neurify,sherlock,fastlin,barrier,barrier-revisited,sdp,star,ai2,bcrown,singh2019boosting,cnn-cert,deepz,muller2022prima,elboher2020abstraction,ryou2021scalable,vegas}.  A key component of verification is devising specifications that accurately characterize the expected behavior of a DNN in \emph{realistic deployment settings}.  Designing such specifications is crucial for ensuring that the corresponding formal guarantees are meaningful and practically relevant.

By and large, the DNN verification community has focused on specifications described by simple analytical expressions.  This line of work has resulted in a set of tools which cover specifications such as certifying the robustness of DNNs against norm-bounded perturbations~\cite{deeppoly,marabou,verinet,bcrown}.  However, while such specifications are useful for certain applications, such as protecting against malicious security threats~\cite{biggio2013evasion}, there are many other applications where real-world distribution shifts, such as change in weather conditions, are more relevant, and these often cannot be described via a set of simple equations. While progress has been made toward broadening the range of specifications~\cite{geometric,paterson2021deepcert,mohapatra2019towards,katz2021verification}, it remains a crucial open challenge to narrow the gap between formal specifications and distribution shifts. 

One promising approach for addressing this challenge involves incorporating neural network components in the specifications~\cite{katz2021verification,mirman2021robustness,xie2022neuro}. Such an approach has been used in the past to verify safety properties of neural network controllers~\cite{katz2021verification} as well as robustness against continuous transformations between images~\cite{mirman2021robustness}. More recently, Xie et al.~\cite{xie2022neuro} generalize this approach by proposing a specification language which can be used to specify complex specifications that are otherwise challenging to define. In this paper, we leverage and extend these insights to obtain a \emph{neural-symbolic} (an integration of machine learning and formal reasoning) approach for verifying robustness against real-world distribution shifts. The key idea is to incorporate deep generative models that represent real-world distribution shifts~\cite{robey2020model,wong2020learning,gowal2020achieving,robey2021model} in the formal specification. 

To realize this idea, there remains one important technical challenge: all the previous work~\cite{katz2021verification,mirman2021robustness,xie2022neuro} assumes that both the neural networks being verified and the generative models are piecewise-linear. This assumption is made in order to leverage off-the-shelf neural network verifiers~\cite{katz2021verification,xie2022neuro}, which focus on piecewise-linear activation functions such as ReLU.
However, in practice the majority of  (image) generative models~\cite{gan,mirza2014conditional,kingma2013auto,sohn2015learning,huang2018multimodal} use \revise{S-shaped activation functions}
\remove{transcendental activation functions} such as sigmoid \revise{and tanh in the output layer}.  While there are a few existing methods for verifying neural networks with sigmoidal activations~\cite{deeppoly,crown,verinet,ryou2021scalable,prima}, they all rely on a one-shot abstraction of the sigmoidal activations and suffer from lack of further progress if the verification fails on the abstraction. To bridge this gap and enable the use of a broad range of generative model architectures in the neural symbolic verification approach, we propose a novel abstraction-refinement algorithm for handling transcendental activation functions. We show that this innovation significantly boosts verification precision when compared to existing approaches.

\begin{figure*}
    \centering
    \begin{subfigure}[b]{0.48\textwidth}
        \centering
        \includegraphics[width=0.7\textwidth]{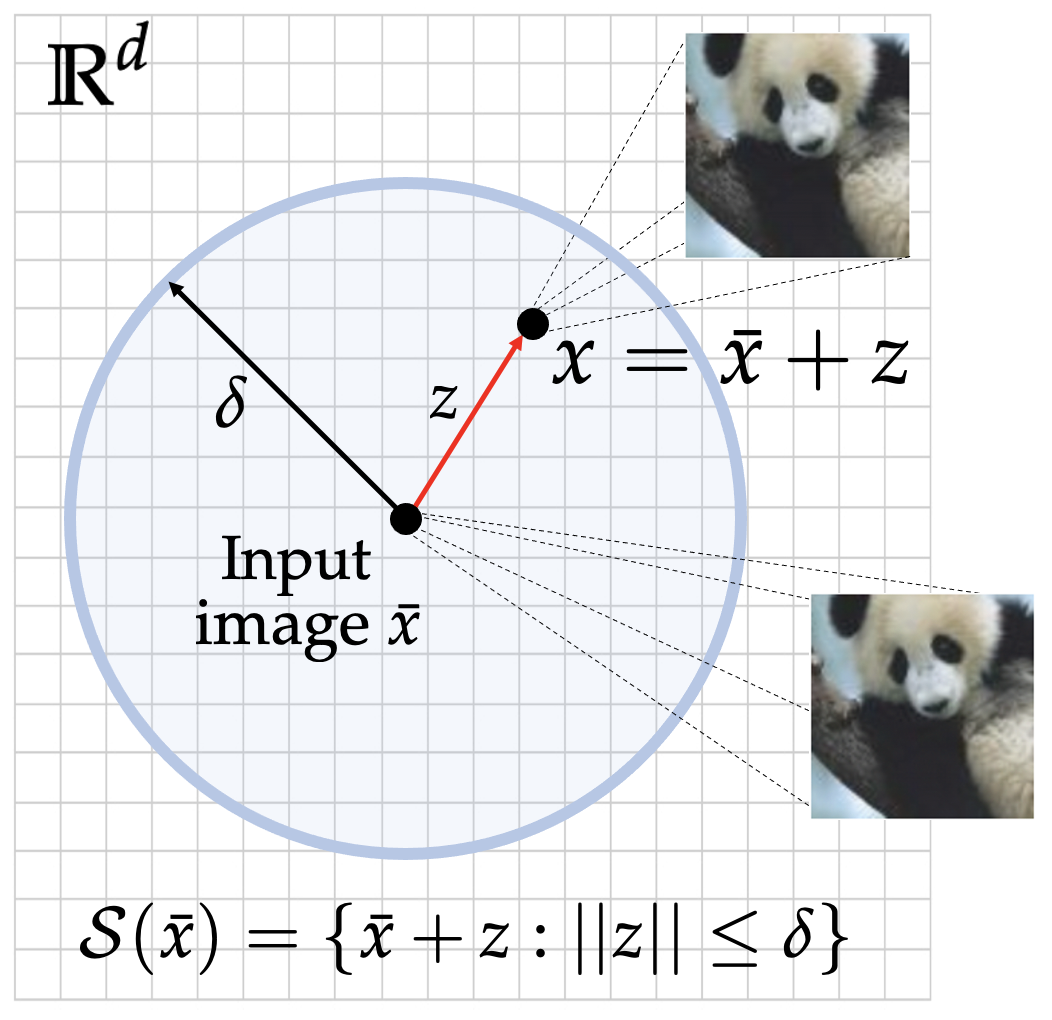}
        \caption{\textbf{Norm-bounded perturbation sets.}  The majority of the verification literature has focused on a limited set of specifications, such as $\ell_p$-norm bounded perturbations, wherein perturbations can be defined by simple analytical expressions.}
        \label{fig:norm-bdd-pert-set}
    \end{subfigure} \hfill
    \begin{subfigure}[b]{0.48\textwidth}
        \centering
        \includegraphics[width=0.8\textwidth]{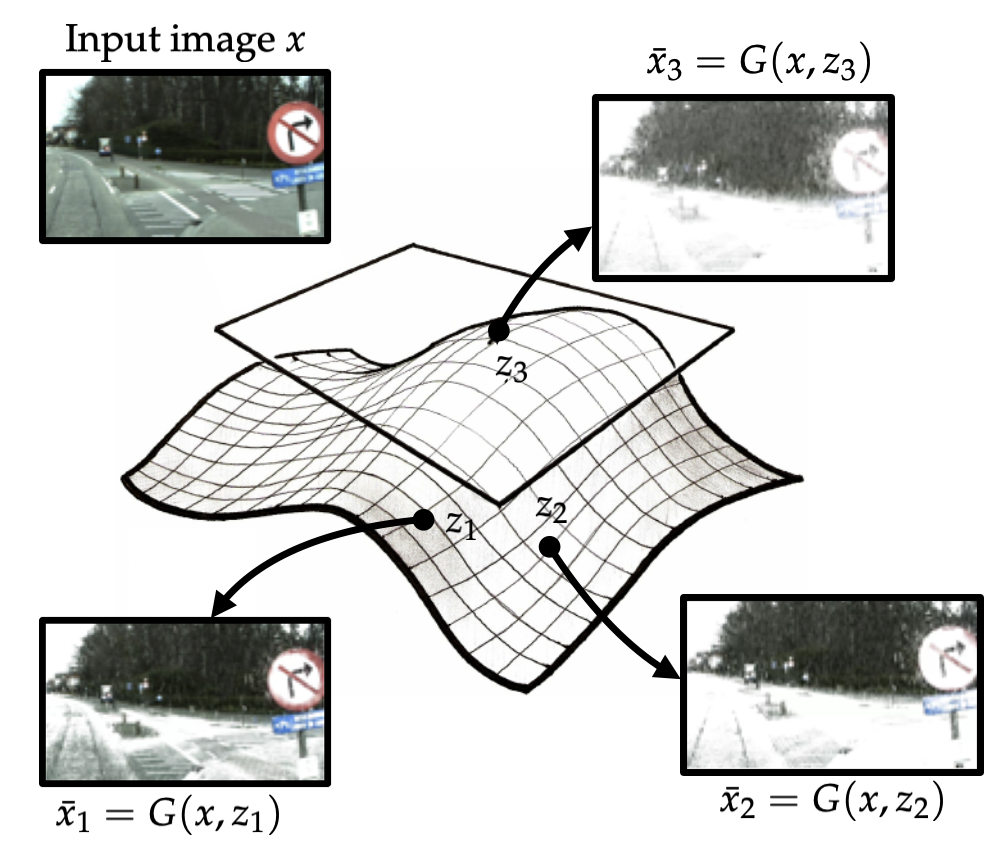}
        \caption{\textbf{Real-world perturbation sets.} Most real-world perturbations cannot be described by simple analytical expressions.  For example, obtaining a simple expression for a perturbation set $\calS(x)$ that describes variation in snow would be challenging.}
        \label{fig:snow-pert-set}
    \end{subfigure}
    \caption{\textbf{Perturbation sets.} We illustrate two examples of perturbation sets $\calS(x)$.}
    \label{fig:pert-sets}
\end{figure*}

Our neural-symbolic approach has the obvious limitation that the quality of the formal guarantee depends on the quality of the neural network used in the specification. However, the approach also possesses several pragmatically useful features that mitigate this limitation: if the verification fails, then a counter-example is produced; by examining the counter-example, we can determine whether it is a failure of the model being tested or a failure of the generative model. In either case, this gives us valuable information to improve the verification framework. \revise{For example, the failure of the generative model might point us to input regions where the generative model is under-trained and data augmentation is needed.} On the other hand, if the verification succeeds, then the generative model represents a large class of inputs for which we know the model is robust.

\vspace{5pt}

\noindent\textbf{Contributions.}  We summarize our contributions are as follows:
\begin{itemize}[nolistsep]
    \item We describe a framework for verifying DNNs against real-world distribution shifts by incorporating deep generative models that capture distribution shifts---e.g.,\ changes in weather conditions or lighting in perception tasks---as verification specifications.
    \item We propose a novel counter-example-guided abstraction refinement strategy for verifying networks with transcendental activation functions.
    \item We show that our verification techniques are significantly more precise than existing techniques on a range of challenging real-world distribution shifts on MNIST and CIFAR-10, as well as on canonical adversarial robustness benchmarks.
\end{itemize}

\section{Problem formulation}
\label{sec:prelim}

In this section, we formally define the problem of verifying the robustness of DNN-based classifiers against real-world distribution shifts.  The key step in our problem formulation is to propose a unification of logical specifications with deep generative models which capture distribution shifts.   

\textbf{Neural network classification.} We consider classification tasks where the data consists of instances $x\in\mathbb{X}\subseteq\R^{d_0}$ and corresponding labels $y \in [k]:=\{1,\dots,k\}$.  The goal of this task is to obtain a classifier $C_f:\R^d\to[k]$ such that $C_f$ can correctly predict the label $y$ of each instance $x$ for each $(x,y)$ pair.  In this work, we consider classifiers $C_f(x)$ defined by
\begin{align}
C_f(x) = \argmax\nolimits_{j\in[k]}\: f_j(x),
\end{align}
where we take $f:\R^{n_0}\to\mathbb{Y}\subseteq \R^{d_L}$ (with $d_L=k$) to be an $L$-layer feed-forward neural network 
with weights and biases $\weight{i} \in \R^{d_i \times d_{i-1}}$ and $\bias{i} \in \R^{d_i}$ for each $i \in [L]$ respectively.  More specifically, we let $f(x) = \layer{L}(x)$ and recursively define
\begin{equation}
     \begin{aligned}
         \layer{i}(x) &= \weight{i}\left(\layerPost{i-1}(x)\right) + \bias{i}, \\ \layerPost{i}({x}) &= \act\left(\layer{i}(x)\right), \quad \text{and} \\
         \layerPost{0}(x) &= x.
    \end{aligned}
\end{equation}
Here, $\act$ is a given activation function (e.g.,\ ReLU, sigmoid, etc.) and $\layer{i}$ and $\layerPost{i}$ represent the pre- and post-activation values of the $i^\text{th}$ layer of $f$ respectively. 

\textbf{Perturbation sets and logical specifications.}  The goal of DNN verification is to determine whether or not a given \emph{logical specification} regarding the behavior of a DNN holds in the classification setting described above. Throughout this work, we use the symbol $\Phi$ to denote such logical specifications, which define relations between the input and output of a DNN.  That is, given input and output properties $\Phi_\text{in}$ and $\Phi_\text{out}$ respectively, we express logical specifications $\Phi$ in the following way:
\begin{align}
    \Phi := (\Phi_{\text{in}}(x) \Rightarrow \Phi_{\text{out}}(y)).
\end{align}
For example, given a fixed instance-label pair $(\bar{x}, \bar{y})$, the specification
\begin{align}
    \Phi := (\norm{\bar{x}- x}_p\leq \epsilon \implies C_f(x) = \bar{y}) \label{eq:norm-bounded-spec}
\end{align}
captures the property of robustness against norm-bounded perturbations by checking whether all points in an $\ell_p$-norm ball centered at $\bar{x}$ are classified by $C_f$ as having the label $\bar y$.

\begin{figure*}
    \begin{subfigure}{0.48\textwidth}
        \centering
        \includegraphics[width=\textwidth]{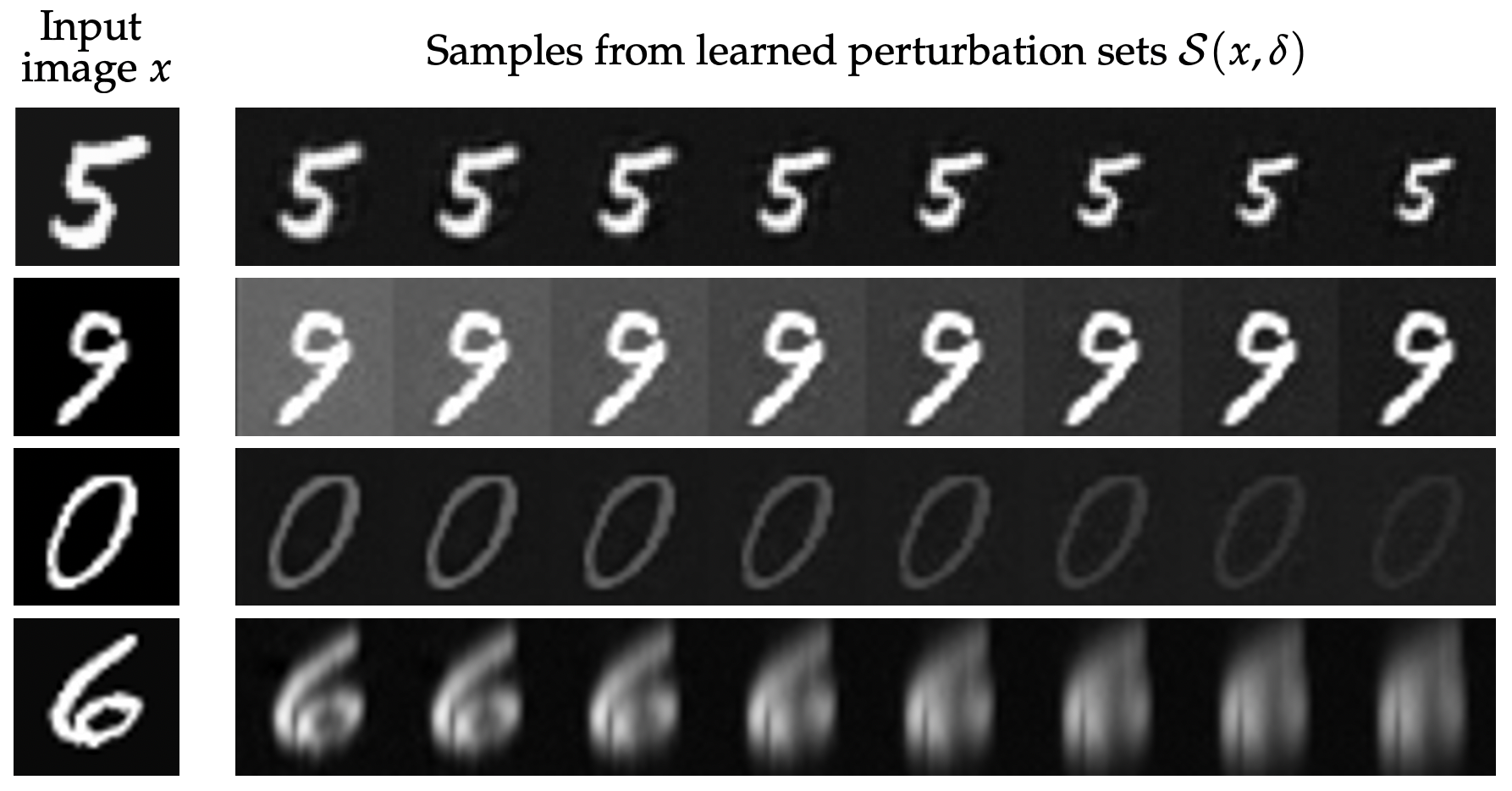}
        \caption{\textbf{MNIST samples.} From top to bottom, the distribution shifts are scale, brightness, contrast, and Gaussian blur.}
        \label{fig:mnist-samples}
    \end{subfigure} \hfill
    \begin{subfigure}{0.48\textwidth}
        \centering
        \includegraphics[width=\textwidth]{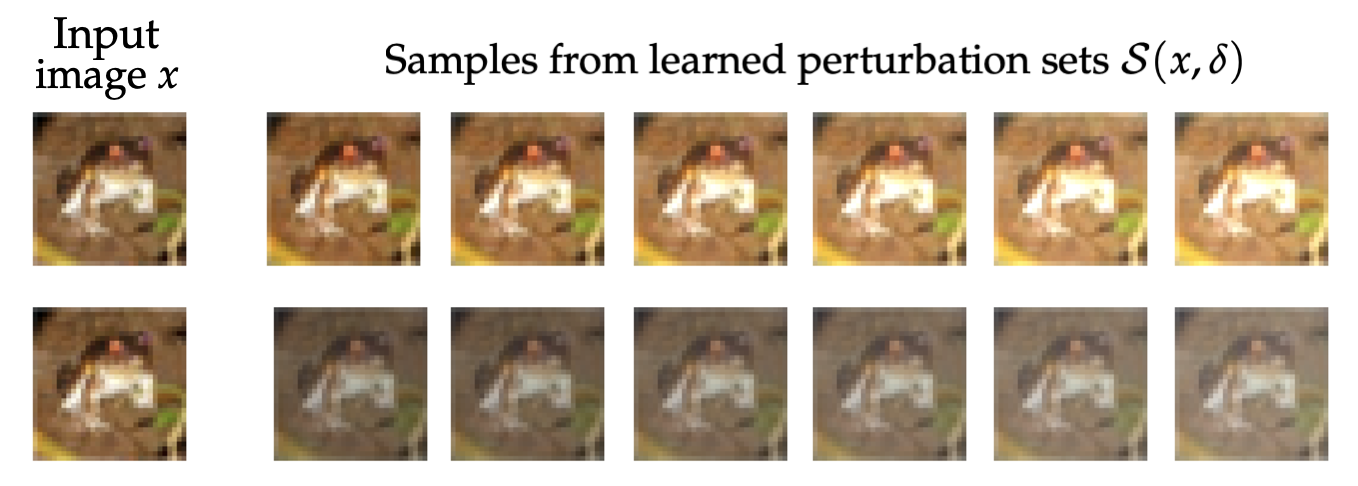}
        \vspace{2em}
        \caption{\textbf{CIFAR-10 samples.} The distribution shifts for these sets are brightness (top) and fog (bottom).}
        \label{fig:cifar-samples}
    \end{subfigure}
    \caption{\textbf{Samples from learned perturbation sets.}  We show samples from two learned perturbation sets $\calS(x)$ on the MNIST and CIFAR-10 datasets.  Samples were generated by gridding the 
    latent space of $\calS(x)$.}
    \label{fig:samples}
\end{figure*}

Although the study of specifications such as~\eqref{eq:norm-bounded-spec} has resulted in numerous verification tools, there are many problems which cannot be described by this simple analytical model, including settings where data varies due to distribution shifts.  For this reason, it is of fundamental interest to generalize such specifications to capture more general forms of variation in data.  To do so, we consider abstract \emph{perturbation sets} $\calS(x)$, which following~\cite{wong2020learning} are defined as ``a set of instances that are considered to be equivalent to [a fixed instance] $x$.''  An example of an abstract perturbation set is illustrated in Figure~\ref{fig:snow-pert-set}, wherein each instance in $\calS(x)$ shows the same street sign with varying levels of snow.  Ultimately, as in the case of norm-bounded robustness, the literature surrounding abstract perturbation sets has sought to train classifiers to predict the same output for each instance in $\calS(x)$~\cite{robey2020model,wong2020learning,gowal2020achieving}.

\textbf{Learning perturbation sets from data.}  Designing abstract perturbation sets $\calS(x)$ which accurately capture realistic deployment settings is critical for providing meaningful guarantees. Recent advances in the generative modeling community have shown that distribution shifts can be \emph{provably} captured by deep generative models.  The key idea in this line of work is to parameterize perturbation sets $\calS(x)$ in the latent space $\calZ$ of a generative model $G(x, z)$, where $G$ takes as input an instance $x$ and a latent variable $z\in\calZ$.  Prominent among such works is~\cite{wong2020learning}, wherein the authors study the ability of conditional variational autoencoders (CVAEs) to capture shifts such as variation in lighting and weather conditions in images.  In this work, given a CVAE parameterized by $G(x, \mu(x) + z\sigma(x))$, where $\mu(x)$ and $\sigma(x)$ are neural networks, the authors consider abstract perturbation sets of the form
\begin{align}
    \calS(x) := \{G(x, \mu(x) + z\sigma(x)) : \norm{z} \leq \delta \}. \label{eq:learned-pert-set}
\end{align}
\revise{In particular, $\mu(x)$ denotes a neural network that maps each instance $x$ to the mean of a normal distribution over the latent space $\calZ$ conditioned on $x$.  Similarly, $\sigma(x)$ maps $x$ to the standard deviation of this distribution in the latent space.\footnote{This notation is consistent with \cite{wong2020learning}, wherein the authors use the same parameterization for conditional VAEs.} It's noteworthy that in this framework, the perturbation upper bound $\delta$ is scaled by $\sigma(x)$, meaning that different instances $x$ and indeed different models $G$ will engender different relative certifiable radii.} 

Under favorable optimization conditions, the authors of~\cite{wong2020learning} prove that CVAEs satisfy two statistical properties which guarantee that the data belonging to learned perturbation sets in the form of~\eqref{eq:learned-pert-set} produce realistic approximations of the true distribution shift (c.f.\ Assumption 1 and Thms.\ 1 and 2 in~\cite{wong2020learning}).  \revise{In particular, the authors of~\cite{wong2020learning} argue that if a learned perturbation set $\calS(x)$ in the form of~\eqref{eq:learned-pert-set} has been trained such that the population-level loss is bounded by two absolute constants, then $\calS(x)$ well-approximates the true distribution shift in the following sense: with high probability over the latent space, for any clean and perturbed pair $(x,\tilde{x})$ corresponding to a real distribution shift, there exists a latent code $z$ such that $\norm{z}_2\leq \alpha$ and $\norm{G(x, \mu(x)z - \sigma(x)) - \tilde{x}} \leq \beta$ for two small constants $\alpha$ and $\beta$ depending on the CVAE loss.  }  To further verify this theoretical evidence, we show that this framework successfully captures real-world shifts on MNIST and CIFAR-10 in Figure~\ref{fig:samples}.

\textbf{Verifying robustness against learned distribution shifts.}
To bridge the gap between formal verification methods and perturbation sets which accurately capture real-world distribution shifts, our approach in this paper is to incorporate perturbation sets parameterized by deep generative models into verification routines.  We summarize this setting in the following problem statement.

\begin{mdframed}[roundcorner=5pt, backgroundcolor=yellow!8,skipabove=12pt, innertopmargin=2pt]
\begin{problem} \label{prob:verification}
Given a DNN-based classifier $C_f(x)$, a fixed instance-label pair $(\bar{x}, \bar{y})$, and an abstract perturbation set $\calS(\bar{x})$ in the form of~\eqref{eq:learned-pert-set} that captures a real-world distribution shift, our goal is to determine whether the following neural-symbolic specification holds:
\begin{align}
    \Phi := \left( x \in\calS(\bar{x}) \implies C_f(x) = \bar{y} \right) \label{eq:real-world-spec}
\end{align}
\end{problem}
\end{mdframed}
In other words, our goal is to devise methods which verify whether a given classifier $C_f$ outputs the correct label $y$ for each instance in a perturbation set $\calS(x)$ parameterized by a generative model~$G$.

\section{Technical approach and challenges}
\label{sec:challenge}

The high-level idea of our approach is to consider the following equivalent specification to~\eqref{eq:real-world-spec}, wherein we absorb the generative model $G$ into the classifier $C$:
\begin{align}
    \Phi = \left( \norm{z} \leq \delta \implies C_{Q_z}(\bar{x}) = \bar{y} \right)
\end{align}
In this expression, we define 
\begin{align}
    Q_z(x) = (f\/ \circ\/ G)(x, \mu(x) + z\sigma(x))
\end{align}
to be the concatenation of the deep generative model $G$ with the DNN $f$.  While this approach has clear parallels with verification schemes within the norm-bounded robustness literature, there is a \emph{fundamental technical challenge}: state-of-the-art generative models \revise{typically use S-shaped activation functions (e.g., sigmoid, tanh) in the last layer}\remove{heavily rely on sigmoid activations} to produce realistic data; however,
the vast majority of the literature concerning DNN verification considers DNNs that are piece-wise linear functions.
Therefore, existing methods for verification of generative models largely do not apply in this setting~\cite{katz2021verification,mirman2021robustness}.

\begin{figure}
\centering
\includegraphics[width=0.4\textwidth]{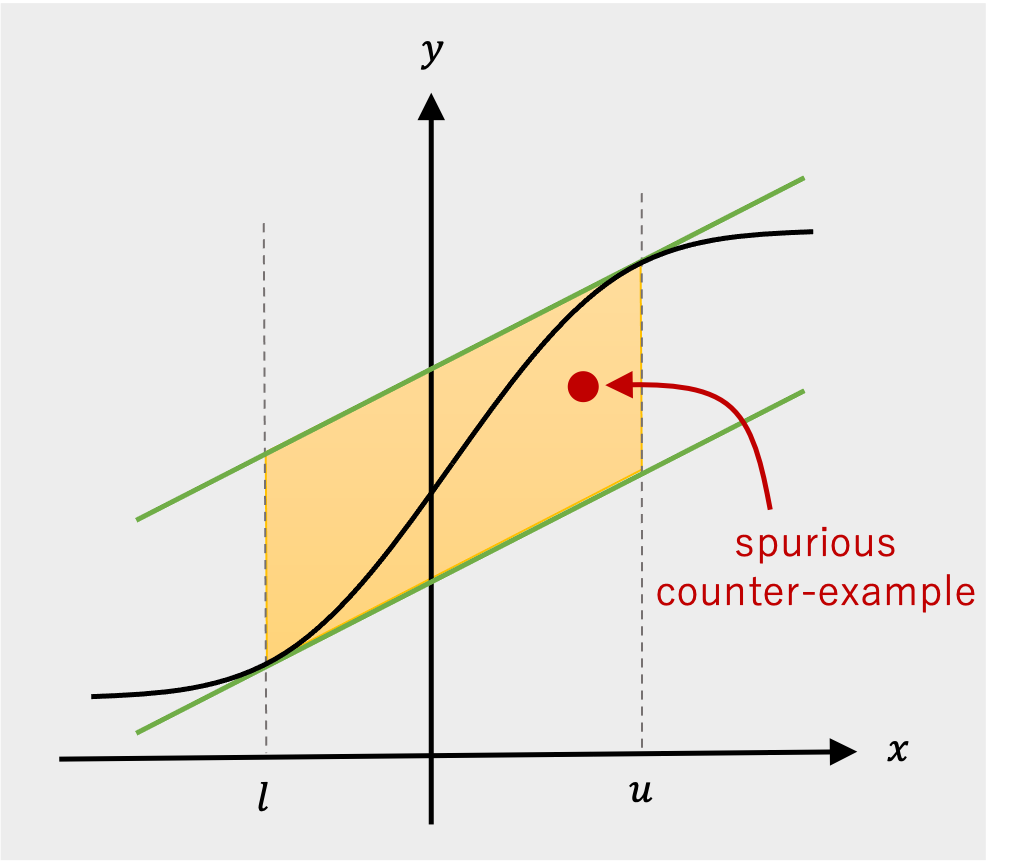}
\caption{An abstraction of the sigmoid activation function.}
\label{fig:deeppoly-sigmoid}
\end{figure}

\textbf{Verification with \revise{S-shaped}\remove{sigmoid} activations.}
In what follows, we describe the challenges inherent to verifying neural networks with \revise{S-shaped}\remove{sigmoid} activations.  For completeness, in the following definitions we provide a formal and general description of \revise{S-shaped}\remove{sigmoid} activations, which will be crucial to our technical approach in this paper.

\begin{definition}[Inflection point] A function $f:\R\to\R$ has an inflection point at $\eta$ iff it is twice differentiable at $\eta$, $f''(\eta) = 0$, and $f'$ changes sign as its argument increases through $\eta$.
\end{definition}
\begin{definition}[\revise{S-shaped}\remove{Sigmoid} function] \label{def:sigmoid}
An \revise{S-shaped}\remove{sigmoid} function $\sig:\R\to\R$ is a bounded, twice differentiable function which has a non-negative derivative at each point and has exactly one inflection point.
\end{definition}

In the wider verification literature, there are a handful of verification techniques that can handle \revise{S-shaped}\remove{sigmoid} functions~\cite{deeppoly,crown,verinet,prima,xu2020fast,xu2020automatic}. \remove{Each of these methods rely on abstraction, which builds an over-approximation of the network behavior. However, abstraction-based methods generally suffer from imprecision depending on the tightness of the abstraction, especially when verifying a neural network on large input domains.}
\revise{Each of these methods is based on \emph{abstraction}, which conservatively approximates the behavior of the neural network. We define abstraction formally in Sec.~\ref{sec:certify}, but illustrate the key ideas here in Fig.~\ref{fig:deeppoly-sigmoid}, which shows the popular sigmoid activation function
\begin{align}
    \rho(x) = \frac{1}{1 + e^{-x}}.
\end{align}
Suppose the input $x$ is bounded between $l$ and $u$, the pre- and post- sigmoid values can be precisely described as $\calD=\{(x,y)\ |\ y = \rho(x)\ \land\ l\leq x \leq u \}$. However, instead of using this precise representation, we could \emph{over-approximate} $\calD$ as $\calD' = \{(x,y)\ |\ y\leq ax + b\ \land\ y \geq cx + d\ \land\ l\leq x \leq u\}$ (the yellow convex region in Fig.~\ref{fig:deeppoly-sigmoid}), where $ax + b$ and $cx + d$ are the two lines crossing the sigmoid function at $(l, \rho(l))$ and $(u, \rho(u))$. $\calD'$ over-approximates $\calD$ because the latter is a subset of the former. Abstraction-based methods typically over-approximate all non-linear connections in the neural network and check whether the specification holds on the over-approximation. The benefit is that reasoning about the (in this case linear) over-approximation is typically computationally easier than reasoning about the concrete representation, and moreover, if the specification holds on the abstraction, then it actually holds on the original network. 

Previous work has studied different ways to over-approximate S-shaped activations, and there are trade-offs between how \emph{precise} the over-approximation is and how \emph{efficient} it is to reason about the over-approximation. For example, a piecewise-linear over-approximation can be more precise than the linear over-approximation (in Fig.~\ref{fig:deeppoly-sigmoid}), but reasoning about the former is more computationally challenging. However, whichever over-approximation one uses, \emph{all} abstraction-based methods suffer from imprecision: if a counter-example to the specification is found on the over-approximation (which means the over-approximation violates the specification), we cannot conclude that the original network also violates the specification. This is because the counter-example may be spurious---inconsistent with the constraints imposed by the precise, unabstracted neural network (as shown in red in Fig.~\ref{fig:deeppoly-sigmoid}).
}

This spurious behavior demonstrates that there is a need for {\em refinement} of abstraction-based methods to improve the precision. For piecewise-linear activations, there is a natural refinement step: performing case analysis on the activation phases. However, when dealing with \revise{S-shaped}\remove{sigmoid} activations, it is less clear how to perform this refinement, because
if the refinement is performed too aggressively, the state space may explode and exceed the capacity of current verifiers. To address this technical challenge, we propose a counter-example guided refinement strategy for \revise{S-shaped}\remove{sigmoid} activation functions which is based on the CEGAR approach~\cite{cegar}.  Our main idea is to limit the scope of the refinement to the region around a specific counter-example.  In the next section, we formally describe our proposed framework and we show that it can be extended to other transcendental activation functions (e.g., softmax).

\section{A CEGAR framework for \revise{S-shaped}\remove{sigmoid} activations}
\label{sec:certify}

\begin{algorithm}[t]
\begin{algorithmic}[1]
\Function{VNN-CEGAR}{$M := \tup{\V, \X, \Y, \linConstraints, \actConstraints}, \Phi$}
\State$M' \assign \abstr(M)$
\While{true}
\State $\tup{\alpha, \textit{proven}} \assign \prove(M', \Phi)$ \Comment{Try to prove.}
\If {\textit{proven}} {\bf return} true \Comment{Property proved.}
\EndIf
\State $\tup{M', \textit{refined}} \assign \refine(M', \Phi, \alpha)$ \Comment{Try to refine.}
\If {$\neg\textit{refined}$} {\bf return} false \Comment{Counter-example is real.}
\EndIf
\EndWhile
\EndFunction
\end{algorithmic}
\caption{VNN-CEGAR($M := \tup{\V, \X, \Y, \linConstraints, \actConstraints}, \Phi$)\label{alg:cegar}}
\end{algorithm}

In this section, we formalize our meta-algorithm for precisely reasoning about DNNs with \revise{S-shaped}\remove{sigmoid} activations, which is based on the CEGAR framework~\cite{mann2021counterexample}. We first present the general framework and then discuss concrete instantiations of the sub-procedures.

\textbf{Verification preliminaries.}  Our procedure operates on tuples of the form $M := \tup{\V, \X, \Y, \linConstraints, \actConstraints}$. Here, $\V$ is a set of real variables with $\X,\Y\subseteq\V$ and $\linConstraints$ and $\actConstraints$ are sets of formulas over $\V$ (when the context is clear, we also use $\linConstraints$ and $\actConstraints$ to mean the conjunctions of the formulas in those sets).
A \emph{variable assignment} $\alpha :\V \mapsto \R$ maps variables in $\V$ to real values. We consider properties of the form $\Phi:= (\Phi_{in}(\X) \Rightarrow \Phi_{out}(\Y))$, where $\Phi_{in}(\X)$ and $\Phi_{out}(\Y)$ are linear arithmetic formulas over $\X$ and $\Y$, and we say that $\Phi$ holds on $M$ if and only if the formula 
$$\psi := \linConstraints \land \actConstraints \land \Phi_{in}(\X) \land \neg \Phi_{out}(\Y)$$
is unsatisfiable. We use $M \models \Phi$ to denote that $\Phi$ holds ($\psi$ is unsatisfiable), $M[\alpha] \models \neg\Phi$ to denote that $\Phi$ does not hold and is falsified by $\alpha$ ($\psi$ can be satisfied with assignment $\alpha$), and $M[\alpha] \models \Phi$ to denote that $\Phi$ is not falsified by $\alpha$ ($\alpha$ does not satisfy $\psi$).  Given this notation, we define a sound abstraction as follows:

\begin{definition}[Sound abstraction] Given a tuple $M:= \tup{\V, \X, \Y, \linConstraints, \actConstraints}$ and a property $\Phi=(\Phi_{in}(X)\Rightarrow \Phi_{out}(Y))$, we say the tuple $M':= \tup{\V'\supseteq\V, \X, \Y, \linConstraintsPrime, \actConstraintsPrime}$ is a sound abstraction of $M$ if $M' \models \Phi$ implies that $M \models \Phi$. 
\label{def:sound-abstraction}
\end{definition}

\textbf{Verifying DNNs.}  Given a DNN $f$, we construct a tuple $M_f$ as follows: for each layer $i$ in $f$, we let $\mathbf{v}^{(i)}$ be a vector of $d_i$ variables representing the pre-activation values in layer $i$, and let $\mathbf{\hat{v}^{(i)}}$ be a similar vector representing the post-activation values in layer $i$.  Let $\mathbf{\hat{v}^{(0)}}$ be a vector of $n_0$ variables from $\V$ representing the inputs.  Then, let $\V$ be the union of all these variables, and let $\X$ and $\Y$ be the input and output variables, respectively; that is, $\X$ consists of the variables in $\mathbf{\hat{v}^{(0)}}$, and $\Y$ contains the variables in $\mathbf{v}^{(L)}$. $\linConstraints$ and $\actConstraints$ capture the affine and non-linear (i.e., activation) transformations in the neural network, respectively. In particular, for each layer $i$, $\linConstraints$ contains the formulas $\mathbf{v}^{(i)} = \weight{i}\mathbf{\hat{v}}^{(i-1)}+\bias{i}$, and $\actConstraints$ contains the formulas $\mathbf{\hat{v}}^{(i)}=\act(\mathbf{v}^{(i)})$.

Algorithm~\ref{alg:cegar} presents a high-level CEGAR loop for checking whether $M \models \Phi$. It is parameterized by three functions. The $\abstr$ function produces an initial \emph{sound abstraction} of $M$. The $\prove$ function checks whether $M' \models \Phi$. If so (i.e., the property $\Phi$ holds for $M'$), it returns with \textit{proven} set to true. Otherwise, it returns an assignment $\alpha$ which constitutes a counter-example. The final function is $\refine$, which takes $M$ and $M'$, the property $P$, and the counterexample $\alpha$ for $M'$ as inputs. Its job is to refine the abstraction until $\alpha$ is no longer a counter-example. If it succeeds, it returns a new sound abstraction $M'$. It fails if $\alpha$ is an actual counter-example for the original $M$. In this case, it sets the return value \textit{refined} to false. Throughout its execution, the algorithm maintains a sound abstraction of $M$ and checks whether the property $\Phi$ holds on the abstraction. If a counter-example $\alpha$ is found such that $M'[\alpha]\models \neg\Phi$, the algorithm uses it to refine the abstraction so that $\alpha$ is no longer a counter-example. The following theorem follows directly from Def.~\ref{def:sound-abstraction}:

\begin{theorem}[CEGAR is sound] Algorithm~\ref{alg:cegar} returns true only if $M \models \Phi$.
\label{theorem:sound-cegar}
\end{theorem}

\subsection{Choice of the underlying verifier and initial abstraction}
The $\prove$ function can be instantiated with an existing DNN verifier. The verifier is required to (1) handle piecewise-linear constraints; and (2) produce counter-examples. 
There are many existing verifiers that meet these requirements~\cite{deeppoly,marabou,verinet,bcrown}. To ensure that these two requirements are sufficient, we also require that $\linConstraintsPrime$ and $\actConstraintsPrime$ only contain linear and piecewise-linear formulas. 

The $\abstr$ function creates an initial abstraction.  For simplicity, we assume that all piecewise-linear formulas are unchanged by the abstraction function.  For \revise{S-shaped}\remove{sigmoid} activations, we use piecewise-linear over-approximations. In principle, any sound piecewise-linear over-approximation of the \revise{S-shaped}\remove{sigmoid} function could be used. One approach is to use a fine-grained over-approximation with piecewise-linear bounds~\cite{overt}. While this approach can arbitrarily reduce over-approximation error, it might easily lead to an explosion of the state space when reasoning about generative models due to the large number of transcendental activations (equal to the dimension of the generated image) present in the system. One key insight of CEGAR is that it is often the case that most of the complexity of the original system is unnecessary for proving the property and eagerly adding it upfront only increases the computational cost. We thus propose starting with a coarse (e.g., convex) over-approximation and only refining with additional piecewise-linear constraints when necessary. Suitable candidates for the initial abstraction of a \revise{S-shaped}\remove{sigmoid} function include the abstraction proposed in \cite{crown,deeppoly,verinet,prima,xu2020fast}, which considers the convex relaxation of the \revise{S-shaped}\remove{sigmoid} activation.

\subsection{Abstraction Refinement for the \revise{S-shaped}\remove{sigmoid} activation function}

\begin{table*}[t]
\vspace{-2mm}
\setlength\tabcolsep{0pt}
\centering
\sffamily
\begin{tabular}{c|c|c|c|c|c}
\toprule
 &
 \begin{minipage}{0.19\textwidth}
\includegraphics[width=\textwidth, height=0.9\textwidth]{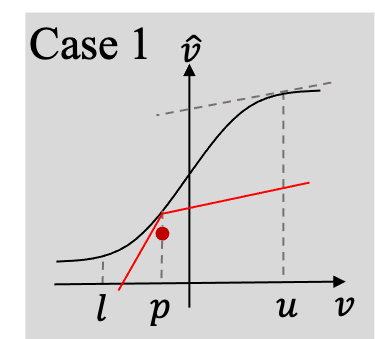}
\end{minipage}& 
\begin{minipage}{0.19\textwidth}
\includegraphics[width=\textwidth, height=0.9\textwidth]{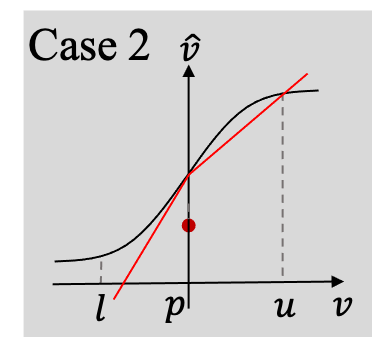}
\end{minipage}
&\begin{minipage}{0.19\textwidth}
\includegraphics[width=\textwidth, height=0.9\textwidth]{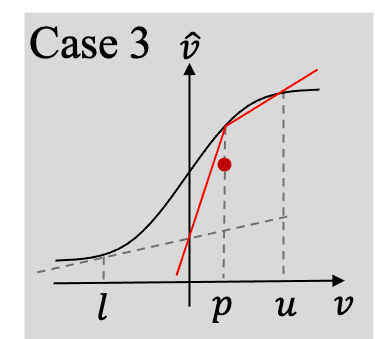}
\end{minipage}
&\begin{minipage}{0.19\textwidth}
\includegraphics[width=\textwidth, height=0.9\textwidth]{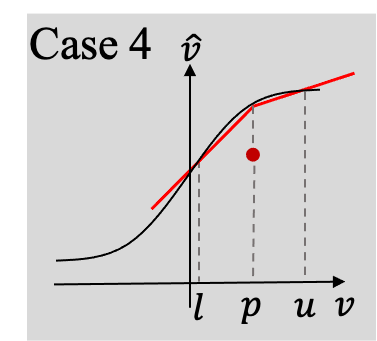}
\end{minipage}
&\begin{minipage}{0.19\textwidth}
\includegraphics[width=\textwidth, height=0.9\textwidth]{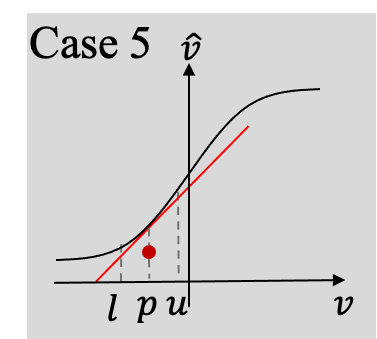}
\end{minipage}\\
\cmidrule(lr){0-5} 
& 
\begin{minipage}{0.19\textwidth}
\centering
$l < \eta, u > \eta$\\ 
$\sig''(p) > 0$ 
\end{minipage}
& 
\begin{minipage}{0.19\textwidth}
\centering
$l < \eta, u > \eta$\\
$\sig''(p) = 0$
\end{minipage}
& 
\begin{minipage}{0.19\textwidth}
\centering
$l < \eta, u > \eta$\\
$\sig''(p) < 0$
\end{minipage}
& 
\begin{minipage}{0.19\textwidth}
\centering
$l > \eta \lor u < \eta$\\
$\sig''(p) \leq 0$
\end{minipage}
& 
\begin{minipage}{0.19\textwidth}
\centering
$l > \eta \lor u < \eta$\\
$\sig''(p) > 0$
\end{minipage}\\
\cmidrule(lr){0-5} 
\begin{minipage}{0.03\textwidth}
$\beta$ 
\end{minipage}
& $\sig'(p)$ 
& $\sig'(p)$
& 
\begin{minipage}{0.19\textwidth}
\centering
$\frac{\sig(l) + \sig'(l) (\eta- l) -\sig(p)}{ \eta - p}$ 
\end{minipage}
& $\frac{\sig(p) - \sig(l)}{p-l}$
& $\sig'(p)$ \\
\cmidrule(lr){0-5} 
\begin{minipage}{0.03\textwidth}
$\gamma$ 
\end{minipage}
& $\min(\sig'(l), \sig'(u))$ 
& $\frac{\sig(p) - \sig(u)}{p-u}$
& $\frac{\sig(p) - \sig(u)}{p-u}$
& $\frac{\sig(p) - \sig(u)}{p-u}$
& $\sig'(p)$ \\
\bottomrule
\end{tabular}
\vspace{2mm}
\caption{Slopes for the piece-wise linear abstraction refinement. \revise{The figures illustrate the refinement on the sigmoid function.}}
\label{tab:slope}
\end{table*}

We now focus on the problem of abstraction refinement for models with \revise{S-shaped}\remove{sigmoid} activation functions. Suppose that an assignment $\alpha$ is found by $\prove$ such that $M'[\alpha] \models \neg\Phi$, but for some neuron with \revise{S-shaped}\remove{sigmoid} activation $\sig$, represented by variables $(v, \hat{v})$, $\alpha(\hat{v}) \neq \sig(\alpha(v))$. The goal is to refine the abstraction $M'$, so that $\alpha$ is no longer a counter-example for the refined model. Here we present a refinement strategy that is applicable to \emph{any} sound abstraction of the \revise{S-shaped}\remove{sigmoid} functions.  We propose using two linear segments to exclude spurious counter-examples.
The key insight is that this is always sufficient for ruling out any counter-example.
We assume that $\linConstraintsPrime$ includes upper and lower bounds for each variable $v$ that is an input to a\revise{n S-shaped}\remove{sigmoid} function. In practice, bounds can be computed with bound-propagation techniques~\cite{reluval,crown,deeppoly}. 

\begin{lemma}
Given an interval $(l,u)$, a\revise{n S-shaped}\remove{sigmoid} function $\sig$, and a point $(p, q) \in \R^2$, where $p\in(l,u)$ and $q \neq \sig(p)$, there exists a piecewise-linear function $\plFun : \R \mapsto \R$ that 1) has two linear segments; 2) evaluates to $\sig(p)$ at $p$; and 3) separates $\set{(p, q)}$ and $\set{(x,y)|x\in(l,u)\wedge y = \sig(x)}$.
\label{lemma:separate}
\end{lemma}
Leveraging this observation, given a point $(p, q) = (\alpha(v), \alpha(\hat{v}))$, we can construct a piecewise-linear function $h$ of the following form:
\begin{equation*}
    h(x) = \sig(p) +
    \begin{cases}
        \beta (x - p) & \text{if } x \leq p\\
        \gamma (x - p) & \text{if } x > p
    \end{cases}
\end{equation*}
that separates the counter-example and the \revise{S-shaped}\remove{sigmoid} function. If $q > \sig(p)$, we add the formula $\hat{v} \leq \plFun(v)$ to the abstraction. And if $q < \sig(p)$, we add $\hat{v} \geq \plFun(v)$. 

The values for the slopes $\beta$ and $\gamma$ should ideally be chosen to minimize the over-approximation error while maintaining soundness. Additionally, they should be easily computable. Table.~\ref{tab:slope} presents a general recipe for choosing $\beta$ and $\gamma$ when the spurious counter-example point is below the \revise{S-shaped}\remove{sigmoid} function. Choosing $\beta$ and $\gamma$ when the counter-example is above the \revise{S-shaped}\remove{sigmoid} function is symmetric (details are shown in App.~\ref{app:slope}). $\eta$ denotes the inflection point of the \revise{S-shaped}\remove{sigmoid} function.

Note that in case 5, $\beta$ is the same as $\gamma$, meaning that a linear bound (the tangent line to $\sig$ at $p$) suffices to exclude the counter-example. In terms of optimality, all but the $\gamma$ value in case 1 and the $\beta$ value in case 3 maximally reduce the over-approximation error among all valid slopes at $(p, \sig(p))$. In those two cases, linear or binary search techniques~\cite{crown,verinet} could be applied to compute better slopes, but the formulas shown give the best approximations we could find without using search.

\begin{lemma}[Soundness of slopes] Choosing $\beta$ and $\gamma$ using the recipe shown in Table~\ref{tab:slope} results in a piecewise-linear function $h$ that satisfies the conditions of Lemma~\ref{lemma:separate}.
\label{thm:sound-slope}
\end{lemma}

\begin{algorithm}[t]
\begin{algorithmic}[1]
\State $\textit{refined} \assign 0$
\For {$(v, \hat{v}) \in \textsf{AllSigmoidal}(\V')$}
\If {$\alpha(\hat{v}) = \sig(\alpha(v))$} {\bf continue} \Comment{Skip satisfied activations.}
\EndIf
\State {$\textit{refined} \assign \textit{refined} + 1$}
\State {$\beta, \gamma \assign \textsf{getSlopes}(l(v), u(v), \alpha(v), \alpha(\hat{v}))$} \Comment{Compute slopes.}
\State {$\actConstraintsPrime \assign \actConstraintsPrime \cup \textsf{ addPLBound}(\beta,\gamma, v, \hat{v}, \alpha)$} \Comment{Refine the abstraction.}
\If {\textsf{stopCondition(\textit{refined})}} {\bf break} \Comment{Check termination condition.}
\EndIf
\EndFor
\State {\bf return} $\tup{\V', \X, \Y, \linConstraintsPrime, \actConstraintsPrime}, \textit{refined} > 0$
\end{algorithmic}
\caption{Refine({$M' := \tup{\V', \X, \Y, \linConstraintsPrime, \actConstraintsPrime}, \Phi, \alpha: \V \mapsto \R$}.\label{alg:refinement}}
\end{algorithm}

An instantiation of the $\refine$ function for neural networks with \revise{S-shaped}\remove{sigmoid} activation is shown in Alg.~\ref{alg:refinement}. It iterates through each \revise{S-shaped}\remove{sigmoid} activation function. For the ones that are violated by the current assignment, the algorithm computes the slopes following the strategy outlined above with the \textsf{getSlopes} function and adds the corresponding piecewise-linear bounds (e.g., $\hat{v} \geq \plFun(v)$) with the \textsf{addPLBound} function. Finally, we also allow the flexibility to terminate the refinement early with a customized \textsf{stopCondition} function. This is likely desirable in practice, as introducing a piecewise-linear bound for each violated activation might be too aggressive.  Furthermore, adding a single piecewise-linear bound already suffices to exclude $\alpha$. We use an adaptive stopping strategy where we allow at most $m$ piecewise-linear bounds to be added in the first invocation of Alg.~\ref{alg:refinement}. And then, in each subsequent round, this number is increased by a factor $k$.  For our evaluation, below, we used $m=30$ and $k=2$, which were the values that performed best in an empirical analysis.

\begin{theorem}[Soundness of refinement] Given a sound abstraction $M'$ of tuple $M$, a property $\Phi$, and a spurious counter-example $\alpha$ s.t.\ $M'[\alpha] \models \neg\Phi$ and $ M[\alpha]\models \Phi$, Alg.~2 produces a sound abstraction of $M$, $M''$, s.t.\ $M''[\alpha] \models \Phi$.
\label{thm:sound-ref}
\end{theorem}

\vspace{-2mm}
\section{Experimental evaluation}
\vspace{-1mm}
\label{sec:experiments}
In this section, we evaluate the performance of our proposed verification framework. 
We begin by investigating the effectiveness of our CEGAR-based approach on boosting the verification accuracy of existing approaches based on one-shot over-approximation. We evaluate on both robustness queries on real-world distribution shifts~(\S\ref{sec:comparison-performance}) and existing benchmarks on robustness against norm-bounded perturbations~\cite{prima,vnncomp} (\S\ref{subsec:adv-robust}). Next, we benchmark the performance of our verifier on a range of challenging distribution shifts (\S~\ref{sec:eval-all-dist-shifts}).  Finally, we use our method to show that robust training tends to result in higher levels of certified robustness against distribution shifts (\S~\ref{sec:verif-robust-training}).

\textbf{Datasets for distribution shifts.}  We consider a diverse array of distribution shifts on the MNIST \cite{lecun1998gradient} and CIFAR-10 \cite{krizhevsky2009learning} datasets.  The code used to generate the perturbations is adapted from \cite{hendrycks2019benchmarking}.

\textbf{Training algorithms.}  For each distribution shift we consider, we train a CVAE using the framework outlined in~\cite{wong2020learning}. \revise{All generators use sigmoid activations in the output layer and ReLU activations in the hidden layers. This is a typical architecture for generative models.} For each dataset, the number of sigmoid activations used in the CVAE is the same as the (flattened) output dimension; that is, 784 ($28\times28$) sigmoids for MNIST and 3072 ($3\times32\times32$) sigmoids for CIFAR-10.  Throughout this section, we use various training algorithms, including empirical risk minimization (ERM)~\cite{vapnik1999nature}, invariant risk minimization (IRM)~\cite{arjovsky2019invariant}, projected gradient descent (PGD)~\cite{pgd}, and model-based dataset augmentation (MDA)~\cite{robey2020model}.  

\textbf{Implementation details.} We use the DeepPoly/CROWN~\cite{deeppoly,crown} method, which over-approximates the sigmoid output with two linear inequalities (an upper and a lower bound), to obtain an initial abstraction (the $\abstr$ function in Alg.~\ref{alg:cegar}) for each sigmoid and instantiate the $\prove$ function with the Marabou neural network verification tool~\cite{marabou}.\footnote{We note that our framework is general and can be used with other abstractions and solvers.
\revise{To motivate more efficient ways to encode and check the refined abstraction, we describe in App.~\ref{app:leaky-encoding} how to encode the added piecewise-linear bounds as LeakyReLUs. This leaves open the possibility of further leveraging neural network verifiers that support LeakyReLUs.}}  
All experiments are run on a cluster equipped with Intel Xeon E5-2637 v4 CPUs running Ubuntu 16.04 with 8 threads and 32GB memory.

\subsection{Evaluation of CEGAR-based verification procedure}
\label{sec:comparison-performance}

\begin{table*}[]
\vspace*{1mm}
\setlength\tabcolsep{4pt}
\centering		
\sffamily
\begin{scriptsize}
\begin{tabular}{lllccccccc}
\toprule
\multirow{3}{*}{\vspace*{2pt}Dataset} & \multirow{3}{*}{Gen.} & \multirow{3}{*}{Class.} & \multicolumn{1}{c}{\deeppoly}
& \multicolumn{2}{c}{\nocegar} & \multicolumn{3}{c}{\cegar} \\
\cmidrule(lr){4-4} \cmidrule(lr){5-6} \cmidrule(lr){7-9} 
& & &  $\delta$ & $\delta$ & time(s) &  $\delta$ & time(s) & \# ref.\     \\
\cmidrule{1-9}
MNIST & $\mgA$ & $\mcA$ & $0.104 \pm 0.041$ & $0.139 \pm 0.058$ & $7.8$ & 
$ \mathbf{0.157} \pm 0.057$ & $84.1$ & $1.5 \pm 1.1$ \\
 & $\mgB$ & $\mcA$ & $0.08 \pm 0.035$ & $0.106 \pm 0.049$ & $20.4$ & $\mathbf{0.118} \pm 0.049$ & $114.8$ & $1.0 \pm 1.1$ \\
 & $\mgA$ & $\mcB$ & $0.102 \pm 0.044$ & $0.136 \pm 0.061$ & $16.4$ & $\mathbf{0.15} \pm 0.059$ & $120.6$ & $1.2 \pm 1.2$ \\
 & $\mgB$ & $\mcB$ & $0.081 \pm 0.037$ & $0.112 \pm 0.049$ & $60.8$ & $\mathbf{0.121} \pm 0.049$ & $191.6$ & $0.8 \pm 1.1$ \\
 & $\mgA$ & $\mcC$ & $0.099 \pm 0.041$ & $0.135 \pm 0.062$ & $41.3$ & $\mathbf{0.146} \pm 0.059$ & $186.9$ & $1.0 \pm 1.1$ \\
 & $\mgB$ & $\mcC$ & $0.082 \pm 0.036$ & $0.116 \pm 0.044$ & $75.7$ & $\mathbf{0.122} \pm 0.041$ & $163.3$ & $0.6 \pm 1.0$ \\
\cmidrule{1-9}
CIFAR & $\cgA$ & $\ccA$ & $0.219 \pm 0.112$ & $0.273 \pm 0.153$ & $33.5$ & $\mathbf{0.287} \pm 0.148$ & $140.8$ & $4.5 \pm 9.2$ \\
 & $\cgB$ & $\ccA$ & $0.131 \pm 0.094$ & $0.18 \pm 0.117$ & $13.7$ & $\mathbf{0.194} \pm 0.115$ & $112.5$ & $3.1 \pm 6.0$ \\
 & $\cgA$ & $\ccB$ & $0.176 \pm 0.108$ & $0.242 \pm 0.14$ & $16.0$ & $\mathbf{0.253} \pm 0.136$ & $57.7$ & $1.6 \pm 2.4$ \\
 & $\cgB$ & $\ccB$ & $0.12 \pm 0.077$ & $0.154 \pm 0.087$ & $7.9$ & $\mathbf{0.172} \pm 0.085$ & $140.2$ & $3.3 \pm 4.2$ \\
\bottomrule
\end{tabular}
\vspace{1mm}
\caption{Evaluation results of three solver configurations. \label{tab:evalcegar}}
\end{scriptsize}
\end{table*}

We first compare the performance of our proposed CEGAR procedure to other baseline verifiers that do not perform abstraction refinement. To do so, we compare the largest perturbation $\delta$ in the latent space of generative models $G$ that each verifier can certify.  In our comparison, we consider three distinct configurations: (1) \deeppoly, which runs the DeepPoly/CROWN abstract interpretation procedure without any search; (2) \nocegar, which runs a branch-and-bound procedure (Marabou) on an encoding where each sigmoid is abstracted with the DeepPoly/CROWN linear bounds and the other parts are precisely encoded; and (3) \cegar, which is the CEGAR method proposed in this work.%
\footnote{We also tried eagerly abstracting the sigmoid with fine-grained piecewise-linear bounds, but the resulting configuration performs much worse than a lazy approach in terms of runtime. Details are shown in App.~\ref{app:eager}} 
For each verifier, we perform a linear search for the largest perturbation bound each configuration can certify.  Specifically, starting from $\delta=0$, we repeatedly increase $\delta$ by $0.02$ and check whether the configuration can prove robustness with respect to $\calS(x)$ within a given time budget (20 minutes). The process terminates when a verifier fails to prove or disprove a given specification.


For this experiment, we consider the \emph{shear} distribution shift on MNIST and the \emph{fog} distribution shift on the CIFAR-10 dataset (see Figure~\ref{fig:samples}).  All classifiers are trained using ERM.  To provide a thorough evaluation, we consider several generator and classifier architectures; details can be found in App.~\ref{app:cvae}.  Our results are enumerated in Table~\ref{tab:evalcegar}, which 
shows the mean and standard deviation of the largest $\delta$ each configuration is able to prove for the first 100 correctly classified test images. We also report the average runtime on the largest $\delta$ proven robust by \nocegar and \cegar, as well as the average number of abstraction refinement rounds by \cegar on those $\delta$ values. Across all configurations, our proposed technique effectively improves the verifiable perturbation bound with moderate runtime overhead.  This suggests that the counter-example guided abstraction refinement scheme can successfully boost the precision when reasoning about sigmoid activations by leveraging existing verifiers. 

\revise{
\subsection{Effect of hyper-parameters $m$ and $k$}
\label{sec:mk_studies}

The refinement procedure (Alg.~\ref{alg:refinement}) has two numerical hyper-parameters, $m$ and $k$, which together control the number of new bounds introduced in a given refinement round (at round $i$, at most $m \times k^i$ new bounds are introduced). If too few new bounds are introduced, then there is not enough refinement, and the number of refinement rounds needed to prove the property increases. On the other hand, if too many bounds are introduced, then the time required by the solver to check the abstraction might unnecessarily increase, resulting in timeouts. 

To study the effect of $m$ and $k$ more closely, we evaluate the runtime performance of \cegar instantiated with different combinations of $m$ and $k$. In particular, we run all combinations of $m \in \{10,30, 50\}$ and $k\in \{1.5, 2\}$ on the first 10 verification instances for the first two generator-classifier pairs in Table~\ref{tab:evalcegar}--$(\mgA, \mcA)$ and $(\cgA, \ccA)$. The perturbation bound is the largest $\delta$ proven robust for each instance.

\begin{table}[h]
\vspace*{1mm}
\setlength\tabcolsep{8pt}
\centering		
\sffamily
\begin{scriptsize}
\begin{tabular}{lllccc}
\toprule
Dataset & $m$ & $k$ & \# solved & time(s) & \# ref.\ \\
\cmidrule{1-6}
MNIST & 10 & 1.5 & 10 & 139.5 & 2.1 \\
     & 30 & 1.5 & 8 & 52.4 & 0.9 \\
     & 50 & 1.5 & 7 & 39.8 & 0.9 \\
     & 10 & 2 & 10 & 113.6 & 1.8 \\
     & 30 & 2 & 10 & 121.0 & 1.1 \\
     & 50 & 2 & 9 & 64.6 & 0.9 \\
\cmidrule{1-6}
CIFAR & 10 & 1.5 & 9 & 77.0 & 2.3 \\
     & 30 & 1.5 & 9 & 60.4 & 1.5 \\
     & 50 & 1.5 & 9 & 38.2 & 1.4 \\
     & 10 & 2 & 9 & 55.3 & 2.4 \\
     & 30 & 2 & 10 & 100.7 & 2.2 \\
     & 50 & 2 & 10 & 82.2 & 1.6 \\
\bottomrule
\end{tabular}
\vspace{1mm}
\caption{Effect of $m$ and $k$ on the runtime performance of CEGAR. \label{tab:evalmk}}
\end{scriptsize}
\end{table}

Table~\ref{tab:evalmk} shows the number of solved instances, the average runtime on solved instances, and the average number of refinement rounds on solved instances. For the same value of $k$, the number of refinement rounds on solved instances consistently decreases as $m$ increases. This is expected, because refinements are performed more eagerly as $m$ increases. However, the decrease in the number of refinement rounds does not necessarily imply improvements in performance. For example, $(50,2)$ solves one fewer MNIST benchmark than $(30,2)$. On the other hand, if the strategy is too ``lazy'' (e.g., $m$ is too small), the increased number of refinement rounds can also result in runtime overhead. For example, on the CIFAR10 benchmarks, when $k=1.5$, the average runtime decreases as $m$ increases, even though all three configurations solve the same number of instances.

Overall, this study suggests that the optimal values of $m$ and $k$ vary across different benchmarks, and exploring adaptive heuristics for choosing these hyper-parameters is a promising direction for boosting the runtime performance of the proposed algorithm. 

}

\subsection{Further evaluation of CEGAR on adversarial robustness benchmarks.}\label{subsec:adv-robust}
To better understand the effectiveness of our abstraction-refinement technique on boosting the verification accuracy over one-shot approximation, we consider a different initial abstraction, PRIMA~\cite{prima}, a more recently proposed abstraction that considers convex relaxations over groups of activation functions and is empirically more precise than DeepPoly/CROWN. In particular, we focus on the \emph{same} sigmoid benchmarks used in \cite{prima}. We use the PRIMA implementation in the artifact associated with the paper%
\footnote{\url{https://dl.acm.org/do/10.1145/3462308/full/}} and run a configuration $\cegarPrima$ which is the same as $\cegar$, except that we run PRIMA instead of DeepPoly/CROWN for the initial abstraction. Each job is given 16 threads and a 30 minute
wall-clock time limit. Table~\ref{tab:prima} shows the number of verified instances and the average runtime on verified instances for the two configurations. Our configuration is able to consistently solve more instances with only a moderate increase in average solving time. This suggests that the meta-algorithm that we propose can leverage the tighter bounds derived by existing abstraction-based methods and boost the state of the art in verification accuracy on sigmoid-based networks.

\begin{table}[h]
\centering
\sffamily
\caption{Comparison with PRIMA~\cite{prima} on the same benchmarks used in ~\cite{prima}} \label{tab:prima}
\begin{scriptsize}
 \begin{tabular}{cccccccc}
\toprule
\multirow{3}{*}{\vspace*{2pt}Model} 
& \multirow{3}{*}{\vspace*{2pt}Acc.} 
& \multirow{3}{*}{\vspace*{2pt}$\epsilon$} & \multicolumn{2}{c}{\texttt{PRIMA}}
& \multicolumn{2}{c}{\cegarPrima} \\
\cmidrule(lr){4-5} \cmidrule(lr){6-7}
& & & robust & time(s) & robust & time(s) \\
\cmidrule{1-7}
6x100    & 99 & 0.015& 52 & 106.5 & \textbf{65} & 119.5 \\ 
9x100    & 99 & 0.015& 57 & 136.0 & \textbf{96} & 323.7 \\ 
6x200    & 99 & 0.012& 65 & 197.9 & \textbf{75} & 260.7 \\ 
ConvSmall& 99 & 0.014 & 56 & 100.5 & \textbf{63} & 157.8 \\ 
\bottomrule
\end{tabular}
\end{scriptsize}
\end{table}

We have also evaluated our techniques on the sigmoid benchmarks used in VNN-COMP 2021 (there are no sigomid benchmarks in VNN-COMP 2022), and we are also able to boost the verification precision over other SoTA solvers such as $\alpha-\beta$-CROWN and VeriNet~\cite{verinet} with our approach. Details can be found in App.~\ref{app:vnn-comp}.


\subsection{Benchmarking our approach on an array of real-world distribution shifts} \label{sec:eval-all-dist-shifts}

\begin{figure}
\begin{minipage}[t]{7.8cm} 
\setlength\tabcolsep{3pt}
\centering		
\sffamily
\begin{scriptsize}
\begin{tabular}{llccccccccc}
\toprule
\multirow{3}{*}{\vspace*{2pt}Dataset} & \multirow{3}{*}{\vspace*{2pt}Perturbation} & \multicolumn{2}{c}{$\delta = 0.1$}
& \multicolumn{2}{c}{$\delta = 0.2$} & \multicolumn{2}{c}{$\delta = 0.5$} \\
\cmidrule(lr){3-4} \cmidrule(lr){5-6} \cmidrule(lr){7-8} 
& & robust & time(s) & robust & time(s) & robust & time(s)    \\
\cmidrule{1-8}
 \multirow{8}{*}{\vspace*{8pt}MNIST} & brightness& 99 & 3.4& 96 & 5.0& 89 & 13.7 \\ 
& rotation& 51 & 38.6& 11 & 80.1& 1 & 177.9 \\ 
& gaussian-blur& 86 &4.7& 79 & 10.8& 65 & 36.5 \\ 
& shear& 76 & 21.4& 4 & 102.6& 0 & 135.6 \\ 
& contrast& 90 & 5.9& 85 & 11.1& 74 & 51.0 \\ 
& scale& 95 & 8.0& 84 & 30.8& 3 & 122.7 \\  

\cmidrule{1-8}
\multirow{5}{*}{\vspace*{8pt}CIFAR10} & brightness& 97 & 3.2 & 96 & 5.2 & 86 & 18.5  \\ 
& contrast& 97 & 3.0 & 95 & 4.6 & 77 & 40.0  \\ 
& fog& 84 & 34.3 & 64 & 69.1 & 11 & 256.0  \\ 
& gaussian-blur& 100 & 2.9 & 99 & 3 & 94 & 10.7  \\ 
\bottomrule
\end{tabular}
\caption{Robustness of ERM against different perturbations. \label{tab:runtime}}
\end{scriptsize}
\end{minipage}
\end{figure}

We next use our proposed verification procedure to evaluate the robustness of classifiers trained using ERM against a wide range of distribution shifts. We select the first 100 correctly classified test images from the respective dataset for each perturbation and verify the robustness of the classifier against the perturbation set. Three values of the perturbation variable $\delta$ are considered: 0.1, 0.2, and 0.5. The architectures we consider for MNIST are $\mgB$ and $\mcC$. For CIFAR-10 we use $\cgB$ and $\ccB$. The verification results are shown in Figure~\ref{tab:runtime}. The ``robust'' columns show the number of instances that our verification procedure is able to certify within a 20 minute timeout. As one would expect, the robustness of each classifier deteriorates as the perturbation budget $\delta$ increases. For instance, for the shear transformation, the classifier is robust on 76 out of the 100 instances when $\delta=0.1$, but is only certified robust on 4 instances when $\delta$ increases to 0.2. Information like this could help system developers to identify perturbation classes for which the network is especially vulnerable and potentially retrain the network accordingly. 

\subsection{Verification for various robust training algorithms} \label{sec:verif-robust-training}

\begin{figure}
\begin{minipage}[t]{\linewidth} 
\centering
\setlength\tabcolsep{1.9pt}
\sffamily
\scriptsize
\begin{tabular}{ccccccc} 
\toprule
\multirow{3}{*}{\vspace*{1.9pt}Dataset} & \multirow{3}{*}{\vspace*{1.9pt}Train.~Alg.} & \multicolumn{3}{c}{Test set Accuracy \%} & \multicolumn{2}{c}{Certified Robust \%}\\
\cmidrule(lr){3-5}\cmidrule(lr){6-7}
&\multicolumn{1}{c}{} & Standard & Generative &  & $\delta=0.05$ & $\delta=0.1$ \\
\cmidrule(lr){1-7}
\multirow{4}{*}{\vspace*{1.9pt}MNIST}
&ERM & 97.9 & 71.6 & & 73.2 & 62.4\\
&IRM & 97.8 & 78.7 &  & 91.4 & 37.0 \\ 
&PGD & 97.0 & 79.5 & & 91.0 & 73.8  \\
&MDA & 97.2 & 96.5 &  & 97.2 & 86.6 \\ 
\bottomrule
\end{tabular}
\caption{Test set accuracy and verification accuracy \label{tab:compare_classifier}}
\vspace{-6mm}
\end{minipage}
\end{figure}

Finally, we compare the robustness and accuracy of classifiers trained using ERM, IRM, PGD, and MDA against the shear distribution shifts on the MNIST dataset.  To this end, we measure the accuracy on the entire test set under the learned perturbation generative models. For each classifier, we then select the first 500 correctly classified images in its dataset and verify the targeted robustness of the classifier against the perturbation. The architectures we use are $\mgB$ and $\mcC$. 

Accuracy and robustness results are presented in Figure~\ref{tab:compare_classifier}. Interestingly, MDA, which is perturbation-aware, outperforms the other three perturbation-agnostic training methods, on both test accuracy and robustness, suggesting that knowing what type of perturbation to anticipate is highly useful. Notice that accuracy on the perturbation set is not necessarily a good proxy for robustness: while the IRM-trained classifier has similar accuracy as the PGD-trained classifier, the former is significantly less robust on the perturbation set with $\delta=0.1$. This further supports the need for including formal verification in the systematic evaluation of neural networks and training algorithms.
\section{Related Work}
\label{sec:related}

\textbf{Beyond norm-bounded perturbations.}  While the literature concerning DNN verification has predominantly focused on robustness against norm-bounded perturbations,
some work has considered other forms of robustness, e.g.,\ against geometric transformations of data \cite{geometric,paterson2021deepcert,mohapatra2019towards}.  However, the perturbations considered are hand-crafted and can be analytically defined by simple models. In contrast, our goal is to verify against real-world distribution shifts that are defined via the output set of a generative model.  Our approach also complements recent work which has sought to incorporate neural symbolic components into formal specifications~\cite{xie2022neuro}. \revise{Our work differs from \cite{xie2022neuro} in two ways. Firstly, Xie et al.\ use classifiers and regressors in their neural specifications, while we use generative models to express perturbation sets in our specifications. Secondly, while Xie et al.\ make black-box use of existing verifiers, we propose a new verification scheme for analyzing networks with sigmoid activations. In general, we believe the idea of leveraging neural components for specification is very promising and should be explored further.}

\textbf{Existing verification approaches.}  Existing DNN verification algorithms broadly fall into one of two categories: search-based methods~\cite{KaBaDiJuKo17Reluplex,marabou,mipverify,scaling,babsr,xu2020fast,planet,dependency,optAndAbs,peregrinn,mip01,dual,dual1,dual2,soi,ferrari2022complete,nnv,dlv} and abstraction-based methods~\cite{deeppoly,kpoly,frown,crown,reluval,neurify,sherlock,fastlin,barrier,barrier-revisited,sdp,star,ai2,bcrown,singh2019boosting,cnn-cert,deepz,muller2022prima,elboher2020abstraction,ryou2021scalable,vegas}.  While several existing solvers can handle sigmoid activation functions~\cite{deeppoly,verinet,crown,prima,xu2020fast}, they rely on one-shot abstraction and lack a refinement scheme for continuous progress. On the other hand, a separate line of work has shown that verifying DNNs containing a single layer of logistic activations is decidable~\cite{ivanov2019verisig}, but the decision procedure proposed in this work is computationally prohibitive.   To overcome these limitations, we propose a meta-algorithm inspired by counter-example-guided abstraction refinement~\cite{cegar} that leverages existing verifiers to solve increasingly refined abstractions. We notice a concurrent work~\cite{zhang2022provably} on formal reasoning of sigmoidal activations which is made available a week before the deadline. The technique is orthogonal to our approach as it again performs one-shot over-approximation of sigmoidal activations.

\revise{CEGAR~\cite{cegar,cimatti2018incremental} is a well-known technique. Our contribution in this area lies in investigating the choices of its parameters (i.e., $\abstr$, $\prove$, and $\refine$) when it comes to verifying neural networks with S-shaped activations.}

\textbf{Verification against distribution shifts.}  The authors of~\cite{wong2020learning} also considered the task of evaluating DNN robustness to real-world distribution shifts; in particular, the approach used in~\cite{wong2020learning} relies on randomized smoothing~\cite{cohen2019certified}.  This scheme provides \emph{probabilistic} guarantees on robustness, whereas our approach (as well as the aforementioned approaches) provides \emph{deterministic} guarantees.  In a separate line of work, several authors have sought to perform verification of deep generative models~\cite{katz2021verification,mirman2021robustness}.  However, each of these works assumes that generative models are piece-wise linear functions, which precludes the use of state-of-the-art models. 





\section{Conclusion}
\label{sec:conclusion}
In this paper, we presented a framework for certifying robustness against real-world distribution shifts.  We proposed using provably trained deep generative models to define formal specifications and a new abstraction-refinement algorithm for verifying them.  Experiments show that our method can certify against larger perturbation sets than previous techniques.

\textbf{Limitations.}  We now discuss some limitations of our framework.   First, like many verification tools, the classifier architectures that our approach can verify are smaller than popular architectures such as ResNet~\cite{he2016deep} and DenseNet~\cite{iandola2014densenet}.  This stems from the limited  capacity of existing DNN verification tools for piecewise-linear functions, which we invoke in the CEGAR loop. Given the rapid growth of the DNN verification community, we are optimistic that the scalability of verifiers will continue to grow rapidly, enabling their use on larger and larger networks.
Another limitation is that the quality of our neural symbolic specification is determined by how well the generative model captures real-world distribution shifts. The mismatch between formal specification and reality is in fact common (and often unavoidable) in formal verification.  And while~\cite{wong2020learning} shows that under favorable conditions, CVAEs can capture distribution shifts, these assumptions may not hold in practice.  For this reason, we envision that in addition to these theoretical results, a necessary future direction will be to involve humans in the verification loop to validate the shifts captured by generative models and the produced counterexamples. This resembles how verification teams work closely with product teams to continually re-evaluate and adjust the specifications in existing industrial settings. \revise{In general, we believe that closing this verification loop (verify, debug, verify again, etc.) is a very interesting future research direction. Finally, when applying our techniques in real industrial settings, another challenge is that collecting training data corresponding to new distribution shifts may be costly (e.g., collecting the same street view under different times of day). However, this cost may be justified in safety-critical domains.}



\bibliographystyle{IEEEtran}
\bibliography{bibli}

\newpage
\appendices
\section{Choices of slopes (Cont.)} \label{app:slope}

We present in Table~\ref{tab:slope2} the general recipe for choosing $\beta$ and $\gamma$ in the case when the violation point is above the \revise{S-shaped}\remove{sigmoid} function.

\begin{table*}[t]
\setlength\tabcolsep{0pt}
\centering
\sffamily
\begin{tabular}{c|c|c|c|c|c}
\toprule
 &
 \begin{minipage}{0.19\textwidth}
\includegraphics[width=\textwidth, height=0.9\textwidth]{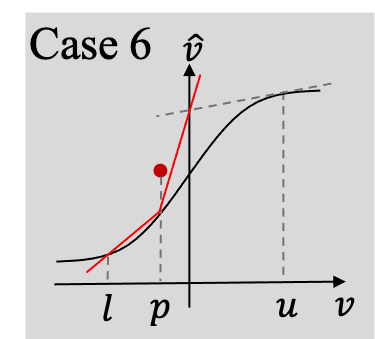}
\end{minipage}& 
\begin{minipage}{0.19\textwidth}
\includegraphics[width=\textwidth, height=0.9\textwidth]{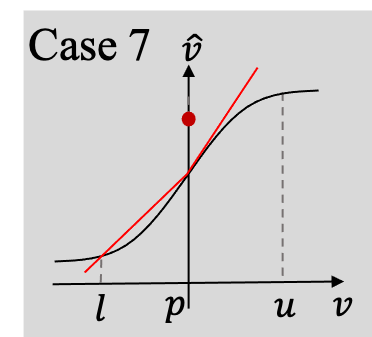}
\end{minipage}
&\begin{minipage}{0.19\textwidth}
\includegraphics[width=\textwidth, height=0.9\textwidth]{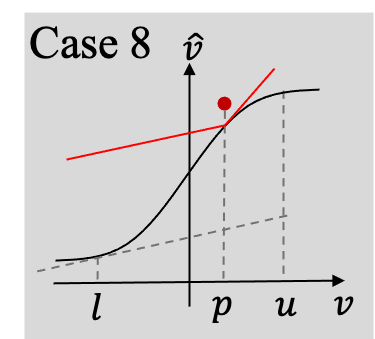}
\end{minipage}
&\begin{minipage}{0.19\textwidth}
\includegraphics[width=\textwidth, height=0.9\textwidth]{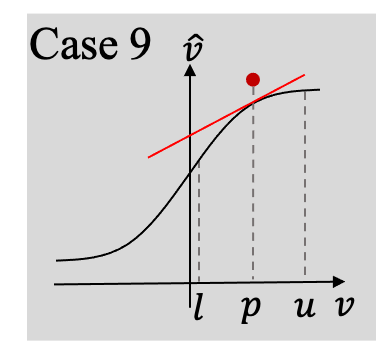}
\end{minipage}
&\begin{minipage}{0.19\textwidth}
\includegraphics[width=\textwidth, height=0.9\textwidth]{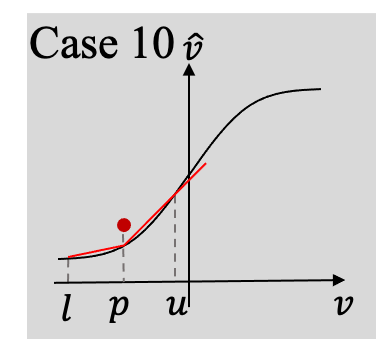}
\end{minipage}\\
\cmidrule(lr){0-5} 
& 
\begin{minipage}{0.19\textwidth}
\centering
$l < \eta, u > \eta$\\ 
$\sig''(p) > 0$ 
\end{minipage}
& 
\begin{minipage}{0.19\textwidth}
\centering
$l < \eta, u > \eta$\\
$\sig''(p) = 0$
\end{minipage}
& 
\begin{minipage}{0.19\textwidth}
\centering
$l < \eta, u > \eta$\\
$\sig''(p) < 0$
\end{minipage}
& 
\begin{minipage}{0.19\textwidth}
\centering
$l > \eta \lor u < \eta$\\
$\sig''(p) \leq 0$
\end{minipage}
& 
\begin{minipage}{0.19\textwidth}
\centering
$l > \eta \lor u < \eta$\\
$\sig''(p) > 0$
\end{minipage}\\
\cmidrule(lr){0-5} 
\begin{minipage}{0.03\textwidth}
$\beta$ 
\end{minipage}
& $\frac{\sig(p) - \sig(l)}{p-l}$
& $\frac{\sig(p) - \sig(l)}{p-l}$
& $\min(\sig'(l), \sig'(u))$ 
& $\sig'(p)$
& $\frac{\sig(p) - \sig(l)}{p-l}$ \\
\cmidrule(lr){0-5} 
\begin{minipage}{0.03\textwidth}
$\gamma$ 
\end{minipage}
&
\begin{minipage}{0.19\textwidth}
\centering
$\frac{\sig(u) - \sig'(u)(u-\eta) -\sig(p)}{ \eta - p}$ 
\end{minipage}
& $\sig'(p)$
& $\sig'(p)$
& $\sig'(p)$
& $\frac{\sig(p) - \sig(u)}{p-u}$ \\
\bottomrule
\end{tabular}

\caption{Slopes for the piece-wise linear abstraction refinement.}
\label{tab:slope2}
\end{table*}

\section{Proofs \label{app:proofs}}

\begin{proof}
\sloppy
\textbf{Theorem~\ref{theorem:sound-cegar}}. Alg.~\ref{alg:cegar} returns true only if the property holds on a sound abstraction of $M$, which following Def.~\ref{def:sound-abstraction} means the property holds on $M$.
\end{proof}

\begin{proof}
\sloppy
\textbf{Lemma~\ref{lemma:separate}}. This can be proved by construction using the $\beta$ and $\gamma$ values in Table~\ref{tab:slope} and Table~\ref{tab:slope2}. We next prove that those choices are sound in Lemma~\ref{thm:sound-slope}.
\end{proof}

Before proving Lemma~\ref{thm:sound-slope}, we first state the following definitions and facts.~\footnote{These are partially adapted from~\cite{cimatti2018incremental}.}

\begin{definition}[Tagent line] The tangent line at $a$ to the function $f$, denoted by $\tangent_{f,a}(x)$, is defined as:  $\tangent_{f,a}(x) = f (a) + f'(a) * (x - a)$.
\end{definition}
\begin{definition}[Secant line]
Definition 2.2. Given $a, b\in \R$, the secant line at $[a,b]$ to a function $f$, denoted by $\secant_{f,a,b}(x)$,
is defined as: $\secant_{f,a,b}(x) = \frac{f(a)-f (b)}{a-b}*(x-a) + f(a)$.
\end{definition}

\begin{proposition}\label{prop:tanBelow}
Let f be a twice differentiable univariate function. If $f''(x) \geq 0$ for all $x \in [l,u]$, then for all $a, x \in
[l,u]$, $\tangent_{f,a}(x)\leq f(x)$, and for all $a,b,x \in [l,u]$, where $a < b$ and $a \leq x \leq b$, $\secant_{f,a,b}(x) \geq f(x)$.
\end{proposition}

\begin{proposition} \label{prop:secBelow}
Let f be a twice differentiable univariate function. If $f''(x) \leq 0$ for all $x \in [l,u]$, then for all $a, x \in
[l,u]$, $\tangent_{f,a}(x)\geq f(x)$, and for all $a,b,x \in [l,u]$, where $a < b$ and $a \leq x \leq b$, $\secant_{f,a,b}(x) \leq f(x)$.
\end{proposition}

\begin{proposition} \label{prop:montonousLeft}
Let f be a univariate function, differentiable with non-negative derivative on $[l, u]$. If $\gamma \leq f'(x)$ for all $x\in [l, u]$, then $f(l) + \gamma (x - l)\leq f(x)$ for all $x \in [l,u]$.
\end{proposition}

\begin{proof}
\sloppy
\textbf{Lemma~\ref{thm:sound-slope}}. Cond.~1 and Cond.~2 hold trivially. Since $q < h(p)$, for Cond.~3, it suffices to show that whenever $x\in(l,u)$, $\sig(x) \ge h(x)$. More concretely, we show that (a) $\sig(x) \geq \sig(p) + \beta(x-p)$ for $x \in [l, p]$, and (b) $\sig(x) \geq \sig(p) + \gamma(x-p)$ for $x \in (p, u]$. We prove this is true for each case in Table.~\ref{tab:slope}. 
\begin{itemize}[noitemsep,topsep=0pt]
    \item \textbf{Case 1}: The segment coresponding to $\beta$ is $\tangent_{\sig,p}$, 
    and Cond.~(a) holds by Prop.~\ref{prop:tanBelow}. On the other hand, the choice $\gamma$ is such that $\gamma \leq \sig'(x)$ for all $x\in[p, u]$. Thus, Cond.~(b) holds by Prop.~\ref{prop:montonousLeft}.
    \item \textbf{Case 2}: The segment coresponding to $\beta$ is $\tangent_{\sig,p}$, 
    so Cond.~(a) holds by Prop.~\ref{prop:tanBelow}. 
    The segment corresponding to $\gamma$ is $\secant_{\sig,p,u}$, 
    so Cond.~(b) holds by Prop.~\ref{prop:secBelow}.
    \item \textbf{Case 3}: For Cond.~(a), we further break it into 2 cases: $x \leq \eta$ and $x > \eta$. In the former case, the line $\sig(p) + \beta(x-p)$ is below the line $\sig(l) + \min(\sig'(l), \sig'(u)) (x - l)$, which by Prop.~\ref{prop:montonousLeft} is below $\sig$. When $x > \eta$, $\sig(p) + \beta(x-p)$ is below the secant line $\secant_{\sig,\eta,p}$, which by Prop.~\ref{prop:secBelow} is below $\sig$. On the other hand, the segment corresponding to $\gamma$ is $\secant_{\sig,p,u}$, so Cond.~(b) holds by Prop.~\ref{prop:secBelow}. 

    \item \textbf{Case 4}: The segments are both secant lines, $\secant_{\sig,l,p}$ and $\secant_{\sig,p,u}$, and thus the conditions hold by Prop.~\ref{prop:secBelow}. 
    \item \textbf{Case 5}: The segments are both tangent lines, $\tangent_{\sig,p}$, and thus the conditions hold by Prop.~\ref{prop:tanBelow}. 
\end{itemize}
The proof for the cases shown in Table.~\ref{tab:slope2} is analogous.
\end{proof}

\begin{proof}
\sloppy{\textbf{Theorem~\ref{thm:sound-ref}.}}
We can prove the soundness of $M''$ by induction on the number of invocations of the \textsc{addPLBound} method. If it is never invoked, then $M'' = M'$ which is a sound abstraction. In the inductive case, it follows from Lemma~\ref{thm:sound-slope} that adding an additional piecewise-linear bound does not exclude variable assignments that respect the precise sigmoid function. On the other hand, when $M[\alpha] \models \Phi$, the \textsc{addPLBound} method will be invoked at least once, which precludes $\alpha$ as a counter-example with respect to $M''$. That is, $M''[\alpha] \models \Phi$.
\end{proof}

\section{Encoding piece-wise linear refinement using LeakyReLU \label{app:leaky-encoding}}
We observe that it is possible to encode the piecewise-linear bounds that we add during abstraction refinement using LeakyReLU functions. While we do not leverage this fact in this paper, we lay out the reduction to LeakyReLU in this section to show how future work usingverification tools supporting LeakyReLUs could benefit.

A LeakyReLU $r_\alpha$ is a piecewise linear function with two linear segments:
\[
r_\alpha(x) =
    \begin{cases}
        \alpha \cdot x & \text{if } x \leq 0 \\
       x & \text{if } x > 0
    \end{cases},
\]
where $\alpha \geq 0$ is a hyper-parameter.

Given a piecewise linear function with two linear segments:
\begin{equation*}
    h(x) - \sig(p) =
    \begin{cases}
        \beta (x - p) & \text{if } x \leq p\\
        \gamma (x - p) & \text{if } x > p
    \end{cases}
\end{equation*}

We can rewrite $h$ as the following:
\begin{equation*}
h(x) = \gamma * r_\alpha(x - p) + \sig(p), \text{ where } \alpha := \frac{\beta}{\gamma}
\end{equation*}

Note that the $\alpha$ value is always valid (i.e., $\alpha\geq 0$) because we always choose both $\beta$ and $\gamma$ to be positive. This means that we can potentially encode the piecewise linear bounds as affine and leaky relu layers. For example, the piecewise linear upper bound $y \leq h(x)$ for a sigmoid $y = \sig(x)$ can be encoded as 
\begin{subequations}
\begin{gather}
a_1 = x - p\\
a_2 = r_\alpha(a_1)\\
a_3 = a_2 + \sig(p)\\
y = a_3 + a_4 \\
a_4 \leq 0,
\end{gather}
\end{subequations}
where $a_1, a_2, a_3, a_4$ are fresh auxilliary variables. Eqs.~a) and c) can be modeled by feed-forward layers. Eq.~b) can be modeled by a leaky relu layer. If we treat $a_4$ as an input variable to the neural network, Eq.~d) can be modeled as a residual connection. This suggests that we could, in principle, express the abstraction as an actual piecewise-linear neural network (with bounds on the input variables (e.g., $a_4$), making it possible to leverage verifiers built on top of neural network software platforms such as Tensorflow or Pytorch.

\section{Details on training and CVAEs} \label{app:cvae}
\subsection{Dataset} \label{sec:generate_perturbation}
We consider the well-known MNIST and CIFAR-10 datasets. The MNIST dataset contains $70,000$ grayscale images of handwritten digits with dimensions $28 \times 28$, where we used $60,000$ images for training and held $10,000$ for testing. The CIFAR-10 dataset contains $60,000$ colored images of $10$ classes with dimensions $3 \times 32 \times 32$, where we used $50,000$ images for training and held $10,000$ for testing.

To perturb the images, we adapt the perturbations implemented in \cite{hendrycks2019benchmarking}.%
\footnote{\url{https://github.com/hendrycks/robustness/blob/master/ImageNet-C/create_c/make_cifar_c.py}}
When training and testing the models, we sample images from the dataset and randomly perturb each image with a strength parameter $c$ that is sampled uniformly from the ranges given in Table \ref{tab:perturbation-strength}.
\begin{table}[h!]
\centering
\sffamily
\caption{Perturbation range in the training data} \label{tab:perturbation-strength}
\begin{scriptsize}
\begin{tabular}{lcc}
\toprule
Dataset & Perturbation & Range of $c$ \\ 
\cmidrule(lr){1-3}
\multirow{7}{*}{\vspace*{0pt}MNIST}   & brightness &  $[.0,.5]$  \\
 & rotation &  $[-60, 60]$  \\
 & gaussian blur &  $[1.0,6.0]$ \\
 & shear &  $[0.2, 1.0]$ \\
 & contrast &  $[0.0, 0.4]$ \\
 & translate &  $[1.0, 5.0]$ \\
 & scale &  $[0.5, 0.9]$ \\
\cmidrule(lr){1-3}
\multirow{4}{*}{\vspace*{0pt}CIFAR10} & brightness &  $[.05,.3]$ \\
 & contrast &  $[.15,.75]$ \\
 & fog &  $[.2,1.5], [1.75, 3]$\\
 & gaussian blur & $[.4,1]$\\
\bottomrule
\end{tabular}
\end{scriptsize}
\end{table}

\subsection{Architecture}

On each dataset, we train a conditional variational encoder (CVAE) with three components: prior network, encoder network, and decoder network (generator). We also train a set of classifiers. In this section, we detail the architecture of these networks. The architectures for the MNIST networks are shown in Tables \ref{tab:mnist-prior}--\ref{tab:mnist-clf3}. Those of the CIFAR networks are shown in Tables \ref{tab:cifar-prior}--\ref{tab:cifar-clf2}. The output layers of the generators use sigmoid activation functions. All hidden-layers use ReLU activation functions.

\begin{table}[h]
\begin{minipage}[h]{.24\textwidth}
\setlength\tabcolsep{5pt}
\centering		
\sffamily
\scriptsize
\caption{Prior} \label{tab:mnist-prior}
\begin{tabular}{cc}
\toprule
Type & Parameters/Shape \\
\cmidrule{1-2}
Input & $28\times28$  \\
\cmidrule{1-2}
Dense & $784\times 1$ \\
\cmidrule{1-2}
Dense & $300\times 1$ \\
\cmidrule{1-2}
Dense & $8\times 2$ \\
\bottomrule
\end{tabular}
\end{minipage}
\vspace{2mm}
\begin{minipage}[h]{.24\textwidth}
\setlength\tabcolsep{3pt}
\centering		
\sffamily
\scriptsize
\caption{Encoder} \label{tab:mnist-recog}
\begin{tabular}{cc}
\toprule
Type & Parameters/Shape \\
\cmidrule{1-2}
Input & $28\times28\times2$ \\
\cmidrule{1-2}
Dense & $784\times 1$ \\
\cmidrule{1-2}
Dense & $300\times 1$ \\
\cmidrule{1-2}
Dense & $8\times 2$ \\
\bottomrule
\end{tabular}
\end{minipage}
\vspace{2mm}
\begin{minipage}[h]{.24\textwidth}
\setlength\tabcolsep{3pt}
\centering		
\sffamily
\scriptsize
\caption{$\mgA$} \label{tab:mnist-dec1}
\begin{tabular}{cc}
\toprule
Type & Param./Shape \\
\cmidrule{1-2}
Input & $28\times28 + 8$ \\
\cmidrule{1-2}
Dense & $200\times 1$ \\
\cmidrule{1-2}
Dense & $784\times 1$ \\
\bottomrule
\end{tabular}
\end{minipage}
\vspace{2mm}
\begin{minipage}[h]{.24\textwidth}
\setlength\tabcolsep{5pt}
\centering		
\sffamily
\scriptsize
\caption{$\mgB$} \label{tab:mnist-dec2}
\begin{tabular}{cc}
\toprule
Type & Param./Shape \\
\cmidrule{1-2}
Input & $28\times28 + 8$ \\
\cmidrule{1-2}
Dense & $400\times 1$ \\
\cmidrule{1-2}
Dense & $784\times 1$ \\
\bottomrule
\end{tabular}
\end{minipage}
\vspace{2mm}
\begin{minipage}[h]{.24\textwidth}
\setlength\tabcolsep{7pt}
\centering		
\sffamily
\caption{$\mcA$}
\label{tab:mnist-clf1}
\scriptsize
\begin{tabular}{cc}
\toprule
Type & Parameters/Shape \\
\cmidrule{1-2}
Input & $28\times28$ \\
\cmidrule{1-2}
Dense & $32\times 1$ \\
\cmidrule{1-2}
Dense & $32\times 1$ \\
\cmidrule{1-2}
Dense & $10\times 1$  \\
\bottomrule
\end{tabular}
\end{minipage}
\vspace{2mm}
\begin{minipage}[h]{.24\textwidth}
\setlength\tabcolsep{7pt}
\centering		
\sffamily
\caption{$\mcB$}
\label{tab:mnist-clf2}
\scriptsize
\begin{tabular}{cc}
\toprule
Type & Parameters/Shape \\
\cmidrule{1-2}
Input & $28\times28$ \\
\cmidrule{1-2}
Dense & $64\times 1$ \\
\cmidrule{1-2}
Dense & $32\times 1$ \\
\cmidrule{1-2}
Dense & $10\times 1$  \\
\bottomrule
\end{tabular}
\end{minipage}
\vspace{2mm}
\begin{minipage}[h]{.24\textwidth}
\setlength\tabcolsep{7pt}
\centering		
\sffamily
\caption{$\mcC$}
\label{tab:mnist-clf3}
\scriptsize
\begin{tabular}{cc}
\toprule
Type & Parameters/Shape \\
\cmidrule{1-2}
Input & $28\times28$ \\
\cmidrule{1-2}
Dense & $128\times 1$ \\
\cmidrule{1-2}
Dense & $64\times 1$ \\
\cmidrule{1-2}
Dense & $10\times 1$  \\
\bottomrule
\end{tabular}
\end{minipage}
\end{table}

\begin{table}[h]
\begin{minipage}[h]{.24\textwidth}
\setlength\tabcolsep{7pt}
\centering		
\sffamily
\caption{Prior}
\label{tab:cifar-prior}
\scriptsize
\begin{tabular}{cc}
\toprule
Type & Parameters/Shape \\
\cmidrule{1-2}
Input & $32\times32\times3$ \\
\cmidrule{1-2}
Dense & $3072\times 1$  \\
\cmidrule{1-2}
Dense & $300\times 1$ \\
\cmidrule{1-2}
Dense & $8\times 2$ \\
\bottomrule
\end{tabular}
\end{minipage}
\vspace{2mm}
\begin{minipage}[h]{.24\textwidth}
\setlength\tabcolsep{7pt}
\centering		
\sffamily
\scriptsize
\caption{Encoder}
\label{tab:cifar-recog}
\begin{tabular}{cc}
\toprule
Type & Parameters/Shape \\
\cmidrule{1-2}
Input & $32\times32\times3\times2$ \\
\cmidrule{1-2}
Dense & $3072\times 1$ \\
\cmidrule{1-2}
Dense & $300\times 1$ \\
\cmidrule{1-2}
Dense & $8\times 2$ \\
\bottomrule
\end{tabular}
\end{minipage}
\vspace{2mm}
\begin{minipage}[h]{.24\textwidth}
\setlength\tabcolsep{4.3pt}
\centering		
\sffamily
\scriptsize
\caption{$\cgA$} \label{tab:cifar-dec1}
\begin{tabular}{cc}
\toprule
Type & Param./Shape \\
\cmidrule{1-2}
Input & $32\times32\times3 + 8$ \\
\cmidrule{1-2}
Dense & $32\times32\times4$ \\
\cmidrule{1-2}
Conv & $3$ $1\times 1$ filters, padding $0$\\
\bottomrule
\end{tabular}
\end{minipage}
\vspace{2mm}
\begin{minipage}[h]{.24\textwidth}
\setlength\tabcolsep{4.3pt}
\centering		
\sffamily
\scriptsize
\caption{$\cgB$} \label{tab:cifar-dec2}
\begin{tabular}{cc}
\toprule
Type & Param./Shape \\
\cmidrule{1-2}
Input & $32\times32\times3 + 8$ \\
\cmidrule{1-2}
Dense & $32\times32\times4$ \\
\cmidrule{1-2}
Conv & $3$ $3\times 3$ filters, padding $1$\\
\bottomrule
\end{tabular}
\end{minipage}
\vspace{2mm}
\begin{minipage}[h]{.24\textwidth}
\setlength\tabcolsep{4.3pt}
\centering
\sffamily
\scriptsize
\caption{$\ccA$}
\label{tab:cifar-clf1}
\begin{tabular}{cc}
\toprule
Type & Params./Shape \\
\cmidrule{1-2}
Input & $32\times32\times3$ \\
\cmidrule{1-2}
Conv & $3$ $3\times 3$ filters, stride $3$  \\
\cmidrule{1-2}
Conv & $3$ $2\times 2$ filters, stride $2$  \\
\cmidrule{1-2}
Dense & $25 \times 1$\\
\cmidrule{1-2}
Dense & $10\times 1$\\
\bottomrule
\end{tabular}
\end{minipage}
\vspace{2mm}
\begin{minipage}[h]{.24\textwidth}
\setlength\tabcolsep{4.3pt}
\centering
\sffamily
\scriptsize
\caption{$\ccB$}
\label{tab:cifar-clf2}
\begin{tabular}{cc}
\toprule
Type & Params./Shape \\
\cmidrule{1-2}
Input & $32\times32\times3$ \\
\cmidrule{1-2}
Conv & $3$ $3\times 3$ filters, stride $2$  \\
\cmidrule{1-2}
Conv & $3$ $2\times 2$ filters, stride $2$  \\
\cmidrule{1-2}
Dense & $25 \times 1$\\
\cmidrule{1-2}
Dense & $10\times 1$\\
\bottomrule
\end{tabular}
\end{minipage}
\end{table}

\subsection{Optimization}
We implement our models and training in PyTorch. The CVAE implementation is adapted from that in \cite{wong2020learning}. On both datasets, we trained our CVAE networks for 150 epochs using the ADAM optimizer with a learning rate of $10^{-4}$ and forgetting factors of 0.9 and 0.999. In addition, we applied cosine annealing learning rate scheduling. Similar to \cite{wong}, we increase $\beta$ linearly from $\beta=0$ at epoch $1$ to $\beta=0.01$ at epoch $40$, before keeping $\beta=0.01$ for the remaining epochs. We use a batch size of 256.

The ERM classifiers on the MNIST dataset are trained with the ADAM optimizer with a learning rate of $10^{-3}$ for $20$ epochs. The classifiers for the CIFAR-10 dataset are trained with the ADAM optimizer with learning rate $10^{-3}$ for $200$ epochs. The classifiers in Sec.~\ref{sec:verif-robust-training} are also all trained with the ADAM optimizer with a learning rate of $10^{-3}$ for 20 epochs.  For PGD, we use a step size of $\alpha=0.1$, a perturbation budget of $\epsilon=0.3$, and we use 7 steps of projected gradient ascent.  For IRM, we use a small held-out validation set to select $\lambda\in\{0.1, 1, 10, 100, 1000\}$.  For MDA, we use a step size of $\alpha=0.1$, a perturbation budget of $\epsilon=1.0$, and we use 10 steps of projected gradient ascent.

\subsection{Computing resources}

The classifiers used in Section~\ref{sec:verif-robust-training} were trained using a single NVIDIA RTX 5000 GPU. The other networks were trained using 8 AMD Ryzen 7 2700 Eight-Core Processors.

\section{Evaluation on VNN-COMP-21 benchmarks}\label{app:vnn-comp}
We also evaluate our techniques on the 36 sigmoid benchmarks used in VNN-COMP-2021. We exclude the benchmark where a counter-example can be found using the PGD attack and evaluate on the remaining 35 benchmarks. In particular, we run a sequential portfolio approach where we first attempt to solve the query with $\alpha$-$\beta$-CROWN~\cite{crown,bcrown,xu2020fast} (competition version), and if the problem is not solved, we run $\cegarPrima$. Table~\ref{tab:prima} shows the results. As a point of comparison, we also report the numbers of the top three performing tools~\cite{xu2020fast,bcrown,verinet,deeppoly,prima} during VNN-COMP-21 on these benchmarks. 
\footnote{\url{https://arxiv.org/abs/2109.00498}}
While $\alpha$-$\beta$-CROWN is already able to solve 29 of the 35 benchmarks, with the abstraction refinement scheme, we are able to solve 1 additional benchmark. We note that during the competition, $\alpha$-$\beta$-crown did not exhaust the 5 minute per-instance timeout on any of these benchmarks.\footnote{\url{https://github.com/stanleybak/vnncomp2021_results/blob/main/results_csv/a-b-CROWN.csv}}
This suggests that the solver was not able to make further progress once the analysis is inconclusive on the one-shot abstraction of the sigmoid activations. On the other hand, our technique provides a viable way to make continuous progress if the one-shot verification attempt fails.

\begin{table*}[h!]
\centering
\sffamily
\caption{Comparison on the VNN-COMP-21 benchmarks}\label{tab:vnn-comp}
\begin{scriptsize}
 \begin{tabular}{cccccccccccc}
\toprule
\multirow{3}{*}{\vspace*{2pt}Model} 
& \multirow{3}{*}{\vspace*{2pt}\# Bench.}
& \multicolumn{2}{c}{$\alpha$-$\beta$-CROWN}
& \multicolumn{2}{c}{VeriNet}
& \multicolumn{2}{c}{ERAN}
& \multicolumn{2}{c}{Ours} \\
\cmidrule(lr){3-4} \cmidrule(lr){5-6} \cmidrule(lr){7-8} \cmidrule(lr){9-10}
& & robust & time(s) & robust & time(s)& robust & time(s) & robust & time(s) \\
\cmidrule{1-10}
6x200 & 35 & 29 & 12.9 & 20 & 2.5 & 19 & 145.5 & \textbf{30}  & 83.2  \\ 
\bottomrule
\end{tabular}
\end{scriptsize}
\end{table*}

\section{Evaluation of an eager refinement strategy}\label{app:eager}

We also compare the lazy abstraction refinement strategy with an eager approach where piecewise-linear bounds are added for each sigmoid from the beginning instead of added lazily as guided by counter-examples. In particular, we attempt to add one piecewise-linear upper-bound and one piecewise-linear lower-bound, each with $K$ linear segments, for each sigmoid activation function. The segment points are evenly distributed along the x-axis. We evaluate on the same MNIST benchmarks as in Table~\ref{tab:evalcegar}, using $K=2$ and $K=3$. The results are shown in Table~\ref{tab:evaleager}. While the two strategies are still able to improve upon the perturbation bounds found by the pure abstract-interpretation-based approach $\deeppoly$, the means of the largest certified $\delta$ values for the eager approach are significantly smaller than those of the CEGAR-based configuration we propose. Interestingly, while $\texttt{K=3}$ uses a finer-grained over-approximation compared with $\texttt{K=2}$, the former only improves on one of the six benchmark sets. This suggests that the finer-grained abstraction increases the overhead to the solver and is not particularly effective at excluding spurious counter-examples on the set of benchmarks that we consider, which supports the need for a more informed abstraction refinement strategy such as the one we propose.

\begin{table*}[t]
\vspace*{1mm}
\setlength\tabcolsep{3pt}
\centering		
\sffamily
\begin{scriptsize}
\begin{tabular}{lllcccccccc}
\toprule
\multirow{3}{*}{\vspace*{2pt}Dataset} & \multirow{3}{*}{Gen.} & \multirow{3}{*}{Class.} & \multicolumn{2}{c}{\texttt{K=2}}
& \multicolumn{2}{c}{\texttt{K=3}} & \multicolumn{3}{c}{\cegar} \\
\cmidrule(lr){4-5} \cmidrule(lr){6-7} \cmidrule(lr){8-10} 
& & &  $\delta$ & time(s) & $\delta$ & time(s) &  $\delta$ & time(s) & \# ref.\     \\
\cmidrule{1-10}
MNIST 
 & $\mgA$ & $\mcA$ & $0.137 \pm 0.043$ & $88.9$ & $0.137\pm0.042$ & $109.8$ & $ \mathbf{0.157} \pm 0.057$ & $84.1$ & $1.5 \pm 1.1$ \\
 & $\mgB$ & $\mcA$ & $0.109 \pm 0.031$ & $114.5$ & $0.109\pm 0.031$ & $199.0$ & $\mathbf{0.118} \pm 0.049$ & $114.8$ & $1.0 \pm 1.1$ \\
 & $\mgA$ & $\mcB$ & $0.126\pm0.045$ & $64.0$ & $0.129	\pm 0.044$ & $95.9$ & $\mathbf{0.15} \pm 0.059$ & $120.6$ & $1.2 \pm 1.2$ \\
 & $\mgB$ & $\mcB$ & $0.108\pm0.038$ &$159.0$ & $0.106\pm0.036$ & $133.8$ & $\mathbf{0.121} \pm 0.049$ & $191.6$ & $0.8 \pm 1.1$ \\
 & $\mgA$ & $\mcC$ & $0.132 \pm 0.043$  & $139.5$ & $ 0.131\pm0.042$ & $190.0$ & $\mathbf{0.146} \pm 0.059$ & $186.9$ & $1.0 \pm 1.1$ \\
 & $\mgB$ & $\mcC$ & $ 0.105\pm	0.033$ & $107.1$ &  $ 0.098\pm0.035$ & $87.5$ & $\mathbf{0.122} \pm 0.041$ & $163.3$ & $0.6 \pm 1.0$ \\
\bottomrule
\end{tabular}
\vspace{1mm}
\caption{Evaluation results of the eager approach. We also report again the results of $\cegar$, which is the same as Table~\ref{tab:evalcegar}. \label{tab:evaleager}}
\vspace{-6mm}
\end{scriptsize}
\end{table*}

\section{Licenses}\label{app:license}

The MNIST and CIFAR-10 datasets are under The MIT License (MIT). The Marabou verification tool is under the terms of the modified BSD license (\url{https://github.com/NeuralNetworkVerification/Marabou/blob/master/COPYING}).

\end{document}